\documentclass[10pt,twocolumn,letterpaper]{article}

\usepackage[pagenumbers]{cvpr} 

\usepackage{graphicx}       
\usepackage{amssymb}        
\usepackage{amsmath}        
\usepackage{mathtools}      
\usepackage{nccmath} 
\usepackage{amsthm}         
\usepackage{lmodern}        
\usepackage{url}            
\usepackage{float}          
\usepackage{ragged2e}       
\usepackage{wrapfig}        
\usepackage{lipsum}         
\usepackage{algorithm}      
\usepackage{algpseudocode}  
\usepackage{threeparttable} 
\usepackage{listings}       

\usepackage{booktabs}       
\usepackage{multirow}       
\usepackage{makecell}       
\usepackage[table]{xcolor}  

\usepackage[font=small]{caption}        
\usepackage[font=footnotesize]{subcaption} 

\usepackage{adjustbox}      
\usepackage[accsupp]{axessibility} 
\usepackage{adjustbox}      

\definecolor{cvprblue}{rgb}{0.21,0.49,0.74}
\definecolor{background}{rgb}{0.98,0.98,0.98}
\definecolor{commentcolor}{rgb}{0.45,0.55,0.55}
\definecolor{keywordcolor}{rgb}{0.2,0.4,0.7}
\definecolor{stringcolor}{rgb}{0.5,0.2,0.3}
\definecolor{numbercolor}{rgb}{0.5,0.5,0.5}
\definecolor{identifiercolor}{rgb}{0.2,0.2,0.2}

\usepackage[
    pagebackref,
    breaklinks,
    colorlinks,
    citecolor=cvprblue,
    linkcolor=red,
    filecolor=magenta,
    urlcolor=magenta
]{hyperref}

\lstdefinelanguage{PythonTorch}{
    language=Python,
    morekeywords={model, optimizer, F, enumerate, float},
    sensitive=true,
    morecomment=[l]{\#},
    morestring=[b]',
    morestring=[b]",
}

\lstdefinestyle{pytorchstyle}{
    language=PythonTorch,
    backgroundcolor=\color{background},
    commentstyle=\color{commentcolor}\ttfamily\small,
    keywordstyle=\color{keywordcolor}\bfseries\ttfamily\small,
    stringstyle=\color{stringcolor}\ttfamily\small,
    identifierstyle=\color{identifiercolor}\ttfamily\small,
    numberstyle=\tiny\color{numbercolor},
    basicstyle=\ttfamily\small,
    breaklines=true,
    captionpos=b,
    numbers=left,
    numbersep=8pt,
    showstringspaces=false,
    tabsize=4,
    frame=single,
    rulecolor=\color{black},
    xleftmargin=15pt,
    xrightmargin=10pt
}

\title{Chain-of-Trajectories: Unlocking the Intrinsic Generative Optimality of Diffusion Models via Graph-Theoretic Planning}

\author{
    Ping Chen$^{1,2}$, Xiang Liu$^{1,2}$, 
      Xingpeng Zhang$^{4}$, Fei Shen$^{3}$, Xun Gong$^{5}$, \\
     Zhaoxiang Liu$^{1,2^*}$, Zezhou Chen$^{1,2}$,
    Huan Hu$^{1,2}$, Kai Wang$^{1,2}$, Shiguo Lian$^{1,2^*}$\\[2mm]
    $^{1}$ Data Science \& Artificial Intelligence Research Institute, China Unicom \\
    $^{2}$ Unicom Data Intelligence, China Unicom,
    $^{3}$ National University of Singapore,\\
$^{4}$ School of Computer Science and Software Engineering, Southwest Petroleum University\\
$^{5}$ School of Computing and Artificial Intelligence, Southwest Jiaotong University \\
    {\tt\small \{chenp181, liuzx178, liansg\}@chinaunicom.cn}
}

\begin{document}
\maketitle
\begin{abstract}
Diffusion models operate in a reflexive \textbf{System 1} mode, constrained by a fixed, content-agnostic sampling schedule. This rigidity arises from the \textbf{curse of state dimensionality}, where the combinatorial explosion of possible states in the high-dimensional noise manifold renders explicit trajectory planning intractable and leads to systematic computational misallocation. To address this, we introduce \textbf{Chain-of-Trajectories (CoTj)}, a train-free framework enabling \textbf{System 2} deliberative planning. Central to CoTj is \textbf{Diffusion DNA}, a low-dimensional signature that quantifies per-stage denoising difficulty and serves as a proxy for the high-dimensional state space, allowing us to reformulate sampling as graph planning on a directed acyclic graph. Through a \textbf{Predict–Plan–Execute} paradigm, CoTj dynamically allocates computational effort to the most challenging generative phases. Experiments across multiple generative models demonstrate that CoTj discovers context-aware trajectories, improving output quality and stability while reducing redundant computation. This work establishes a new foundation for resource-aware, planning-based diffusion modeling. The code is available at \textcolor{blue}{\url{https://github.com/UnicomAI/CoTj}}.
\end{abstract}

\section{Introduction}

\begin{figure*}[ht] 
  \centering
  \includegraphics[width=\linewidth]{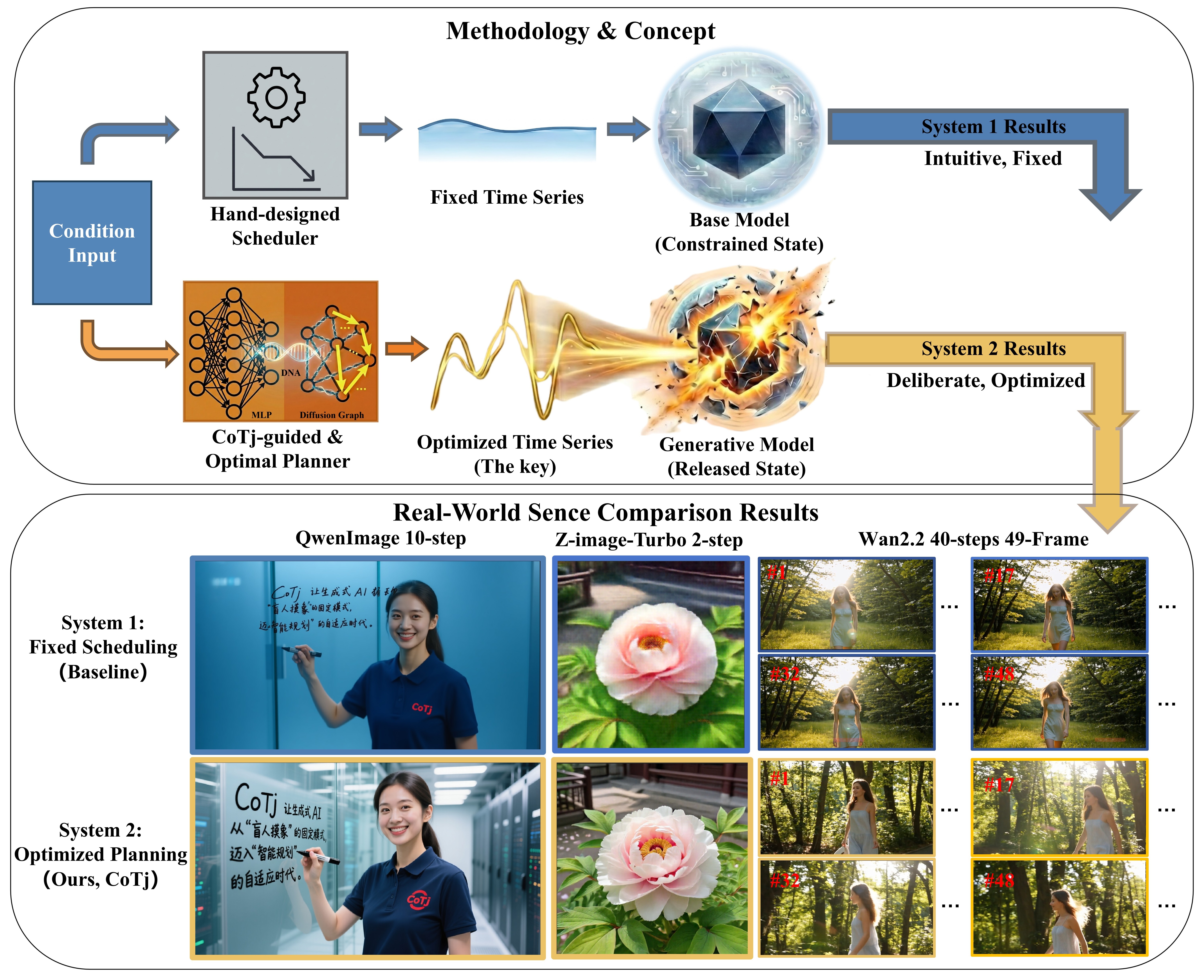} 
  \caption{
  \textbf{The Paradigm Shift from Fixed Scheduling to Chain-of-Trajectories (CoTj).} \textbf{Top:} Comparison of inference mechanisms. Standard diffusion acts as a \textbf{System 1} process, relying on a fixed time series that constrains potential. In contrast, CoTj introduces a \textbf{System 2} approach via a CoTj-guided Optimal Planner, generating an optimized time series tailored to the input. \textbf{Bottom:} Real-world comparisons across modalities. In text-to-image generation (QwenImage \cite{Wu2025QwenImage}, Z-image-Turbo \cite{Cai2025zimage}), CoTj achieves superior visual results under limited computational resources. In text-to-video tasks (Wan2.2 \cite{wan2025wan}), our CoTj significantly enhances motion dynamics and realism while maintaining high spatial quality.
  }
  \label{fig:concept}
\end{figure*}

Diffusion-based generative models have advanced rapidly through iterative denoising. The success of modern diffusion models (DMs) is fundamentally based on the reverse-solving framework of stochastic differential equations (SDEs) or ordinary differential equations (ODEs). The DDPM \cite{Ho2020Denoising} and the fractional base generative model \cite{song2021score} laid the theoretical foundation for them. Subsequent efficiency optimization studies, such as DDIM \cite{Song2021DDIM} and DPM-Solver \cite{Lu2022DPM}, mainly focused on reducing the number of function evaluations (NFE) through higher-order solvers or linear multistep methods, thereby significantly accelerating sampling \cite{chen2025optimizing1}. However, from the perspective of cognitive science, the existing methods generally operate in a ``reflective" (System 1) mode\cite{kahneman2011thinking}: they rely on predefined, content-agnostic sampling scheduling strategies that evenly distribute computing resources throughout the generation process. This rigid approach ignores the inherent heterogeneity (Intrinsic Heterogeneity) in the generation path.

The deeper problem stems from the ``state dimension curse'': the high-dimensional, continuous, and multimodal noise manifold corresponding to the diffusion process and the conditional distribution makes it computationally infeasible to perform explicit and adaptive trajectory search for each specific generated content \cite{karras2022elucidating}. For example, ShortDF \cite{chen2025optimizing1} first recognized this dimensionality challenge and introduced a relaxation-based shortest-path loss to optimize residual propagation in the diffusion process. While effective in improving efficiency and stability, ShortDF's approach remains input-agnostic and does not adapt the sampling trajectory to the semantic content of individual prompts. Consequently, current diffusion models are essentially still limited by the ``execute-as-infer" paradigm, using a single fixed strategy to handle all generation scenarios, unable to dynamically adjust computing resource allocation based on semantic content and generation difficulty, resulting in significant efficiency waste.
Theoretically, adaptive trajectory optimization for specific samples can fundamentally solve the problem of computational resource mismatch in the generation process of diffusion models. However, this approach faces the classic ``dimension disaster'' caused by high-dimensional noise manifolds in practice. As Karras et al. \cite{karras2022elucidating} pointed out, the state space of the diffusion process exhibits combinatorial explosion growth, making traditional search-based planning algorithms in the pixel-level space infeasible and prone to getting stuck in local optima. Although some studies have attempted to adjust the adaptive step size through local error estimation (e.g., the work of Jolicoeur-Martineau et al. \cite{Alexia2021Gotta}), these methods are essentially greedy and local, lacking the ability to conduct a global examination and structured planning of the generation path. This predicament highlights the necessity of mapping the generation process to a low-dimensional, interpretable latent space for representation, which is the key bottleneck in introducing advanced planning mechanisms. In contrast, advances in large language models (LLMs) have shown that explicit intermediate reasoning can substantially enhance complex decision-making. Chain-of-Thought prompting~\cite{Wei2022COT} and the Tree-of-Thought framework~\cite{Yao2023Tree} formalized this idea by decomposing tasks into structured reasoning trajectories that approximate deliberate ``System~2'' cognition. Related efforts have further investigated alternative structured reasoning mechanisms in LLMs~\cite{chen2025fuzzy}. However, this ``prediction-planning-execution'' paradigm has not yet been effectively transferred to the visual generation domain: most existing works are limited to semantic planning at the prompt level rather than global optimization of the underlying numerical calculation path. Current mainstream improvement measures, whether aiming to compress knowledge distillation and consistency models (such as the work of Song et al. \cite{Song2023Consistency} - CM), or enhancing the efficiency along fixed paths with higher-order solvers (such as Lu et al. \cite{lu2025DPM-Solver++} - DPM-Solver++), or based on caching techniques \cite{gao2025lemica,bu2025dicache,liu2025timestep, gao2026meancache,ma2024deepcache}, have not broken through the ``content-independent'' static scheduling paradigm. They replace the understanding of explicit, adaptive trajectories with more efficient static shortcuts, but fail to establish a mechanism that can dynamically reconfigure the global generation path based on the specific semantic content and real-time error characteristics of the input sample. Therefore, diffusion generation is essentially still limited to a rigid ``execution-as-reasoning'' mode, and to move towards true ``System 2'' generation, the core challenge lies in how to construct a planable low-dimensional cognitive map for the high-dimensional continuous visual manifold.

To tackle this fundamental challenge, we introduce \textbf{Chain-of-Trajectories (CoTj)}\textbf{(illustrated in Figure \ref{fig:concept})}, a framework that shifts diffusion sampling from static execution toward \textbf{System 2} style deliberative planning. The core of our approach is \textbf{Diffusion DNA}, a low-dimensional structural signature that captures the distribution of error-correction difficulty across generative stages. Diffusion DNA serves as a computable proxy for the high-dimensional stochastic state space, enabling the construction of a deterministic planning graph. Under formal conditions regarding error bounds and trajectory drift, we reformulate the sampling problem as a global shortest-path optimization on a directed acyclic graph (DAG), where nodes represent intermediate states, edges denote potential temporal jumps, and edge costs are derived from Diffusion DNA. The optimal sampling trajectory is then obtained by solving for the minimum-cost path, which dynamically concentrates computational effort on the most challenging phases of denoising.

Building on this formalism, we propose a \textbf{Predict, Plan, Execute} inference paradigm. First, a lightweight predictor estimates the input-conditioned Diffusion DNA. A planning graph is then constructed, and its optimal path is computed prior to any generative steps. This planning stage requires no retraining of the base diffusion model and incurs negligible overhead. Different inputs yield distinct Diffusion DNA profiles, leading to adaptive, context-aware trajectories. Our experiments reveal that this planning naturally emerges as semantically-aware computation: simple prompts (e.g., ``a dark sky'') yield streamlined, shortcut-heavy paths, while high-entropy, complex descriptions (e.g., ``Van Gogh and Redon style, bright swirling scene'') trigger deliberate, multi-step refinement. This adaptability manifests in two modes: under a fixed computational budget, planning redistributes the denoising sequence according to difficulty; in an unconstrained setting, the trajectory length adjusts automatically to maintain output quality, favoring concise paths for simple content and engaging in extensive refinement for complex generations. Diffusion DNA also provides a structural diagnostic for error dynamics, enabling a comparative analysis of denoising consistency across different base models and exposing hidden instabilities in distilled or few-step variants.

In summary, CoTj transforms diffusion from a rigid, schedule-driven System 1 process into a graph-optimized, planning-guided System 2 framework. By enabling dynamic resource allocation and structured error correction, it advances generative models toward more interpretable and deliberate reasoning. Experiments across image and video generation demonstrate that globally planned trajectories enhance output stability and quality under equivalent computational budgets while reducing redundant steps. This work provides a new theoretical foundation for evolving generative AI from reactive execution toward structured, resource-aware planning.

\section{Related Work}

The evolution of generative diffusion models has largely focused on two orthogonal axes: improving the expressivity of the underlying backbone (e.g., scaling to Transformers and Flow Matching) and accelerating the sampling process. Our work introduces a third axis: inference-time planning. 

\noindent\textbf{From Fixed Schedules to Heuristic Acceleration.}
The standard diffusion sampling paradigm relies on fixed, discretization-agnostic schedules. Early acceleration efforts primarily targeted the mathematical efficiency of the solver itself. Higher-order solvers, such as DPM-Solver \cite{Lu2022DPM} and DEIS \cite{Zhang2023Fast, Blasingame2024AdjointDEIS}, exploit the semi-linear structure of the diffusion ODE to reduce discretization errors, enabling fewer function evaluations (NFEs). Parallel efforts in Knowledge Distillation, including Progressive Distillation \cite{SalimansH2022Progressive}, Consistency Models \cite{Song2023Consistency}, and Rectified Flow \cite{Liu2023Flow,esser2024scaling,yan2024perflow}, attempt to compress the entire trajectory into a single or a few steps.

While effective at reducing latency, these approaches operate under a ``Blind Execution'' paradigm. Distillation methods bake a specific trajectory into the model weights, sacrificing the flexibility to trade compute for quality at inference time. Similarly, advanced solvers optimize how to move between fixed time steps but do not question whether those time steps are the optimal waypoints. Unlike CoTj, which treats the trajectory as a dynamic decision process, these methods view the schedule as a static constraint to be satisfied rather than an objective to be optimized.

\noindent\textbf{Adaptive Sampling and Reactive Error Correction.}
Recognizing that fixed schedules are suboptimal, recent research has explored adaptive sampling. Borrowing from classical ODE integration, methods based on PID control or local error estimation \cite{Zhao2023UniPC} attempt to dynamically adjust step sizes during inference. For instance, approaches like AutoDiffusion \cite{Li2023AutoDiffusion} and Align Your Steps (AYS) \cite{Sabour2024Align} monitor the curvature or Fisher information of the generative vector field, reducing step sizes in regions of high volatility.

However, these methods are fundamentally Reactive (System 1). They act based on local, immediate feedback or dataset-level statistics—detecting an error only after or during a step (or via stochastic corrections like in Score-SDE \cite{song2021score} and EDM \cite{karras2022elucidating}), rather than anticipating the global difficulty landscape. This creates a ``greedy'' optimization behavior that lacks foresight. In contrast, CoTj introduces a Proactive (System 2) mechanism. By predicting the ``Diffusion DNA'' (the global error landscape) before generation begins, our framework solves a global optimization problem, ensuring that computational resources are allocated based on a holistic understanding of the task, akin to solving a boundary-value problem rather than an initial-value problem.

\noindent\textbf{System-2 Reasoning and Planning in Generative AI.}
The distinction between fast, intuitive response (System 1) and slow, deliberative reasoning (System 2) has become a central theme in Large Language Model (LLM) research. Techniques like Chain-of-Thought (CoT) \cite{Wei2022COT}, and Tree of Thoughts \cite{Yao2023Tree} demonstrate that allocating computational budget to intermediate reasoning steps significantly boosts performance on complex tasks.

Despite this success in NLP, visual generation has remained largely dominated by System 1 approaches—executing learned patterns without deliberation. Chain-of-Trajectories bridges this gap. It represents one of the first principled attempts to endow diffusion models with a ``pre-computatio'' planning phase. Unlike iterative refinement methods that add more noise and denoise again (e.g., SDEdit \cite{Meng2022SDEdit}), CoTj performs latent planning: it reasons about the optimal path in the abstract cost space of the DAG before committing to a single pixel update. This aligns with the ``Predict-then-Plan'' paradigm in robotics, separating the estimation of environmental dynamics from the execution of control policies.

\noindent\textbf{Optimal Control and Physics-Inspired Generative Dynamics.}
Theoretically, our formulation resonates with the Principle of Least Action in physics and Optimal Control Theory. Recent works in Flow Matching \cite{lipman2022flow} and Schrödinger Bridges \cite{Somnath2023Aligned} have framed generation as an Optimal Transport problem, seeking paths that minimize kinetic energy or transport cost. However, these formulations typically optimize the continuous vector field during training.

CoTj addresses the complementary discretization control problem at inference time. By mapping the continuous flow onto a discrete graph with costs derived from reconstruction error bounds, we provide a discrete counterpart to the Lagrangian mechanics of the generative process. Our ``Path of Least Action'' is not merely a metaphor but a direct consequence of minimizing the cumulative deviation from the model's canonical manifold, offering a unified geometric interpretation of sampling efficiency that extends beyond heuristic step selection.

\section{Method: Chain-of-Trajectories}

To endow diffusion models with \textbf{System~2 deliberative planning}, we propose \textbf{Chain-of-Trajectories (CoTj)}. The key challenge is the \textbf{curse of state dimensionality}: stochastic diffusion unfolds in a continuous, high-dimensional state space, where every potential step is coupled with noise realizations and semantic variations. Explicit trajectory search is therefore intractable.

\subsection{Diffusion DNA: Task-Conditional Surrogate for Trajectory Planning}
\label{sec:method_dna}



Our key insight is that optimal denoising planning does not require simulating all stochastic outcomes. Instead, for a given generation condition such as a text prompt, control signal, or initialization, we encode expected reconstruction difficulty over time into a low-dimensional \textbf{Diffusion DNA} that deterministically guides trajectory planning.

\noindent\textbf{Postulate 1: The Upper Bound of Correctability.}  
We view denoising as a multi-step corrective integration starting from the canonical noisy state $\mathbf{x}_t^*$ at timestep $t$ along the forward diffusion trajectory. For any meaningful integration path that restores the original data, the initial reconstruction error $\mathcal{C}(t)$ serves as an intrinsic upper bound on the error that can be corrected. We provide a rigorous proof of this error bound in Appendix \ref{sec:postulate1}. Formally, we define the \textbf{Reconstruction Error Reference} as
\begin{equation}
\mathcal{C}(t) \equiv \mathbb{E}_{\mathbf{x}_0, \mathbf{z}}\bigl[\|\hat{\mathbf{x}}_0(\mathbf{x}_t^*, t) - \mathbf{x}_0\|^2\bigr],
\end{equation}
where $\mathbf{x}_0$ denotes the clean sample, and 
$\hat{\mathbf{x}}_0(\mathbf{x}_t^*, t)$ is the model’s single-step reconstruction estimate from the state $\mathbf{x}_t^*$ at timestep $t$. We define the sequence 
\(\mathcal{D} = (\mathcal{C}(0), \dots, \mathcal{C}(T))\) 
as the task-conditional \textbf{Diffusion DNA}, which provides a compact, deterministic surrogate of the high-dimensional diffusion landscape.

\noindent\textbf{Postulate 2: The Canonical Reference Trajectory.}
In practical sampling, the realized state $\mathbf{x}_t$ often drifts from the underlying data manifold due to discretization errors. However, the \textbf{ideal state} $\mathbf{x}_t^* = \alpha_t \mathbf{x}_0 + \sigma_t \mathbf{z}$ perfectly adheres to the marginal distribution $q_t(\mathbf{x})$.
We anchor all cost calculations to this canonical ideal trajectory. As proven in Appendix \ref{sec:postulate2}, this canonical state mathematically guarantees the tightest possible reconstruction error bound; any deviation into an "off-manifold" state strictly incurs an out-of-distribution penalty. This abstraction is crucial: it effectively decouples planning from stochastic realization, converting an intractable path-dependent control problem into a \textbf{tractable deterministic optimization} over time indices.

\noindent\textbf{The Jump-Deviation Trade-off.}  
A solver step from $t$ to $k$ ($t > k$) introduces a drift between the resulting state $\mathbf{x}_k$ and the ideal state $\mathbf{x}_k^*$. Intuitively, larger jumps produce larger drift, as the local curvature of the vector field may change over the interval, causing ``off-manifold'' deviation.  

We quantify this as the \textbf{Trajectory Correction Cost} $W(t,k)$, a weighted projection of the Diffusion DNA:
\begin{equation}
\label{eq:correction_cost}
W(t, k) \approx \|\mathbf{x}_k - \mathbf{x}_k^*\|^2 = s(t, k) \cdot \mathcal{C}(t),
\end{equation}
where $\mathbf{x}_k$ is obtained by propagating $\mathbf{x}_t^*$ to $k$, and $s(t,k)$ is a temporal lever increasing with the jump interval (related to the model's variance schedule; e.g., $s(t,k) = ((t-k)/t)^2$ for linear flow matching, derived in Appendix \ref{sec:stk}).

Importantly, $W(t,k)$ depends only on the source timestep $t$. Once the Diffusion DNA is computed along the canonical trajectory with $T$ evaluations, all transition costs can be determined algebraically, avoiding the combinatorial explosion of stochastic rollouts in the high-dimensional state space. This collapses trajectory optimization from an intractable stochastic search into a tractable deterministic problem over timestep indices, directly overcoming the curse of dimensionality.

The core \textbf{planning trade-off} is thus: small steps ($k \approx t$) incur low immediate cost but consume the error budget slowly, while large jumps ($k \ll t$) reduce remaining error faster at higher instantaneous cost.
\begin{figure}[t]
  \centering
  \includegraphics[width=\linewidth]{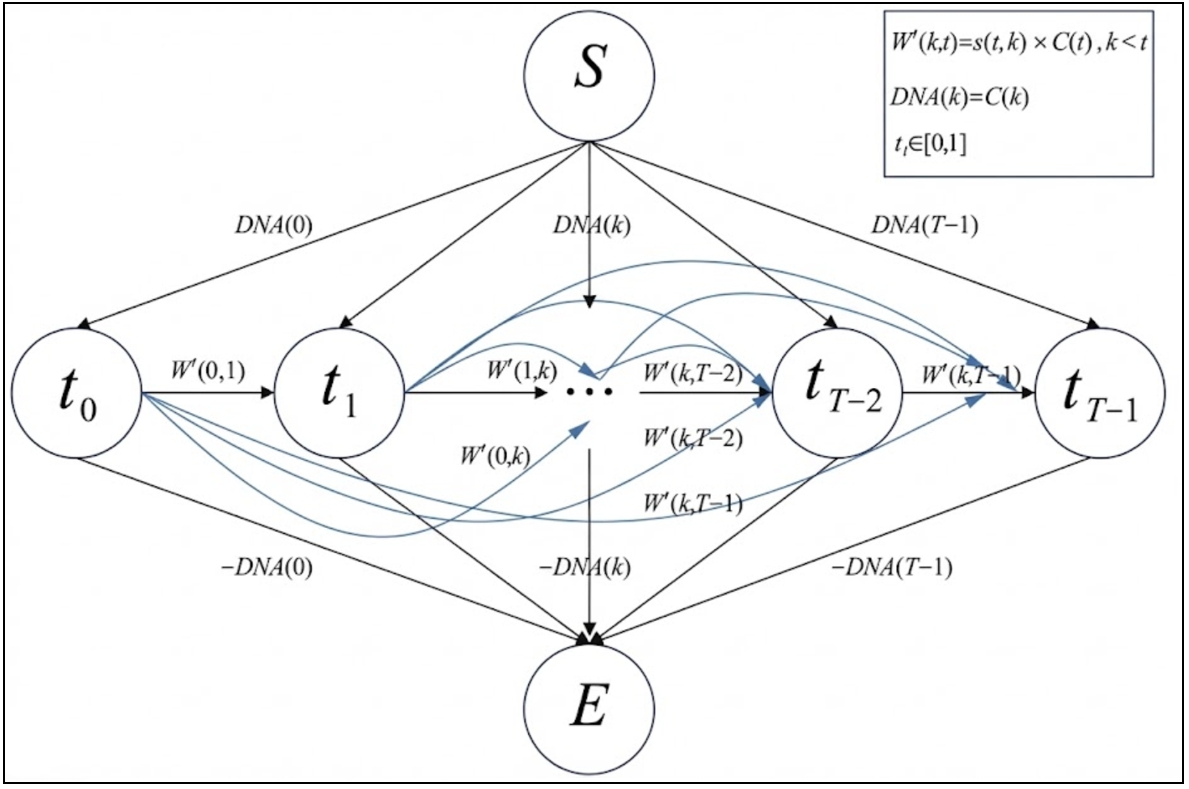}
  \caption{
  \textbf{Super-Node DAG for trajectory planning.} 
  Visualization of the dense reverse-time DAG $\mathcal{G}=(\mathcal{V},\mathcal{E})$, showing aggregated high-dimensional states as Super-Nodes and possible transitions as edges. The DAG encodes all feasible generative trajectories, allowing the CoTj planner to find globally optimal paths while avoiding regressive or non-convergent regions.
  }
  \label{fig:dag}
\end{figure}

\subsection{Graph-Theoretic Planning via Super-DAG}
\label{sec:method_dna2graph}

Leveraging these principles and the On-Manifold Correction Assumption (proven in Appendix \ref{sec:corollary1}), we embed all feasible state transitions into a dense reverse-time directed acyclic graph (DAG), $\mathcal{G}=(\mathcal{V},\mathcal{E})$, adopting a \textbf{Super-Node} topology (Fig.~\ref{fig:dag}). This formulation enables global trajectory optimization by balancing local correction costs against cumulative error reduction, effectively converting high-dimensional stochastic evolution into a structured planning problem.
\begin{itemize}
    \item \textbf{Super-Source ($\mathbf{S}$):} edges $\mathbf{S}\to k$ encode residual error if denoising stops at $k$, $W'(\mathbf{S},k)=\mathcal{C}(k)$ (\emph{Terminal Risk}).
    \item \textbf{Super-End ($\mathbf{E}$):} edges $k \to \mathbf{E}$ encode initial information credit, $W'(k,\mathbf{E})=-\mathcal{C}(k)$.
    \item \textbf{Transition edges ($k\to t$, $t>k$):} weighted by $W'(k,t)=W(t,k)$.
\end{itemize}

The optimal trajectory is obtained via shortest-path search:
\begin{equation}
P^* = \arg\min_{P:\mathbf{S}\to\mathbf{E}} \sum_{(k,t)\in P} W'(k,t),
\end{equation}
balancing local correction cost with global error reduction.  

\emph{Dual interpretation:} This shortest-path formulation is dual to maximizing the \textbf{Global Error Reduction Gain}. By minimizing the accumulated correction costs, the agent effectively selects a path that preserves the highest fidelity relative to the canonical manifold.

\subsection{Fixed-Step and Adaptive Planning}
\label{sec:method_planing}

CoTj naturally supports two operational regimes.  

\textbf{Fixed-Step Budget:} We impose a step budget $|P|=K$, recovering conventional sampling lengths (e.g., 10 steps) while globally optimizing \emph{which} timesteps are selected to minimize total error.

\textbf{Adaptive-Length:}  
Let $W_{\min}$ and $W_{\max}$ denote the minimum total correction cost along the full path of maximum allowed steps and the minimal cost among all single-step jumps, respectively. For a partial trajectory $P_{\text{partial}}$ covering the first $n$ steps, the explained gain ratio is defined as
\begin{equation}
\rho(n) = \frac{W_{\max} - W(P_{\text{partial}})}{W_{\max} - W_{\min}}.
\end{equation}

Under this normalization, $\rho(n)$ can be interpreted as the proportion of denoising improvement achieved relative to the reference-quality trajectory. The adaptive planner terminates sampling once $\rho(n) \ge \rho_{\text{th}}$, where $\rho_{\text{th}}$ is a predefined threshold (e.g., $0.99$), meaning that a specified percentage of the reference-quality improvement has been explained. This design captures most of the attainable denoising gain while avoiding redundant steps. Complex, high-entropy conditions trigger longer, deliberative paths, whereas simpler conditions produce shorter trajectories, reflecting dynamic, System~2-like allocation of computational effort. 

\subsection{Predict-then-Plan: Amortized Deliberation}
\label{sec:ptp}

We observe that, for the same diffusion model, DNA profiles across different conditions exhibit strong structural similarity. To exploit this, we train a lightweight regressor $\phi$ using a cosine similarity loss, encouraging predicted DNA to preserve the structural patterns observed across conditions.

At inference, $\phi$ estimates the DNA from the condition embedding,
\begin{equation}
\hat{\mathcal{D}} = \phi(\mathbf{e}_{\text{cond}}),
\end{equation}
and deterministic graph planning computes the optimal trajectory $P^*$. The diffusion model executes denoising along this path without modification, enabling real-time, adaptive generation.

Inference follows a \textbf{Think-Strategize-Act} loop: predict the DNA, plan the shortest path, and execute denoising. This transforms diffusion from reactive execution into resource-aware generative intelligence with minimal overhead.

\begin{figure}[t]
\centering
\includegraphics[width=\linewidth]{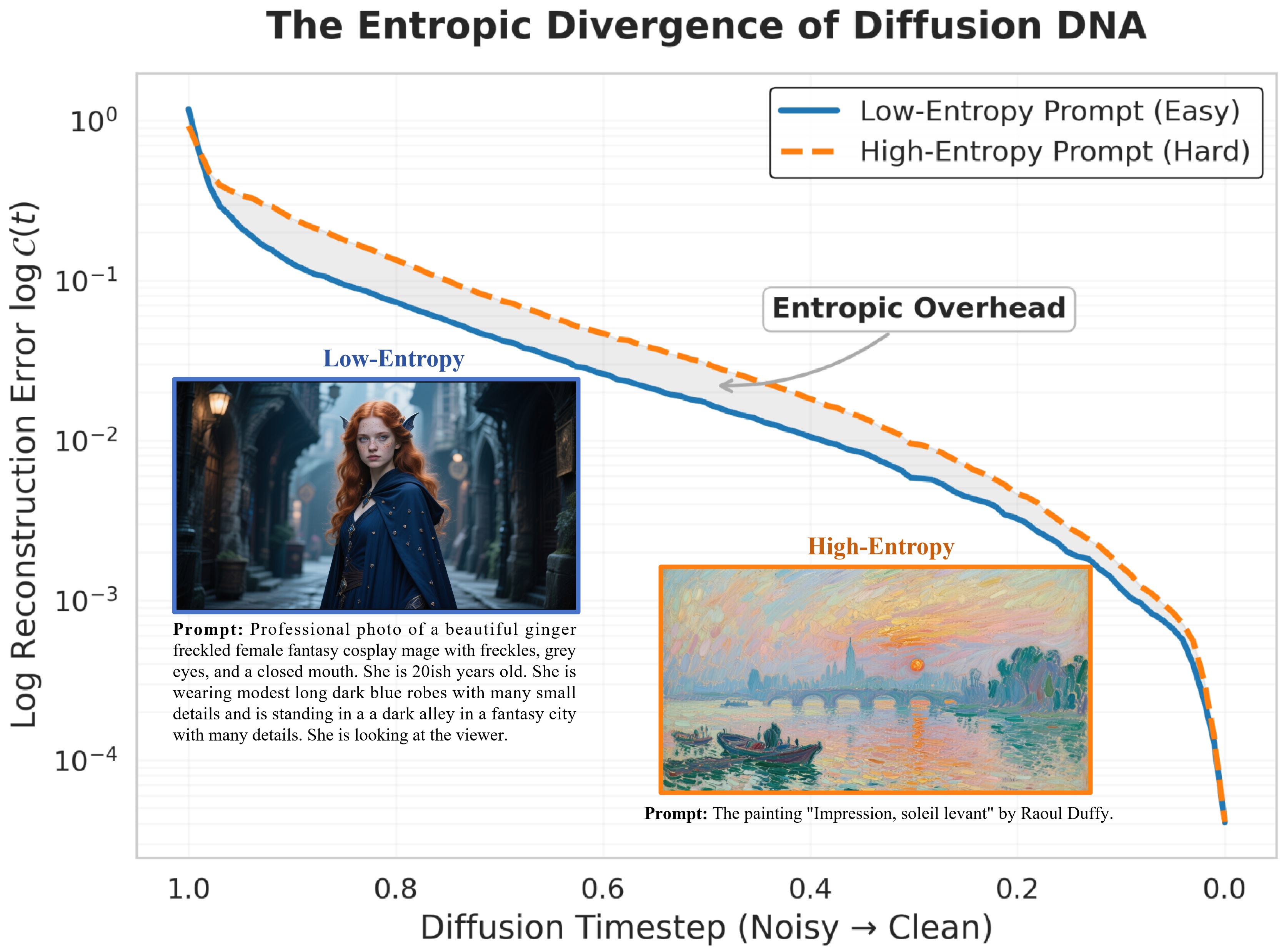} 
\caption{
\textbf{Entropic divergence patterns reveal intrinsic Diffusion DNA.}
Profiles of $\mathcal{C}(t)$ expose heterogeneous error-decay dynamics across inputs.
Structurally constrained, low-entropy visual targets exhibit rapid stabilization,
whereas abstract and visually uncertain compositions lead to prolonged refinement phases.
The shaded region illustrates the additional computational burden induced by
entropic complexity, highlighting that generative difficulty is governed primarily
by intrinsic visual uncertainty rather than surface-level prompt properties
(e.g., length or linguistic complexity).
}

\label{fig:dna_divergence}
\end{figure}

\section{Results}
\subsection{Revealing the Intrinsic Diffusion DNA}
\label{sec:dna_discovery}

Standard diffusion schedules assume uniform denoising dynamics via fixed discretization, yet reconstruction error is inherently tied to noise scale and semantic uncertainty. As a fast, representative model for analysis, we take Qwen-Image \cite{Wu2025QwenImage} and sample 100 uniformly spaced timesteps over the interval $[0, T]$, extracting the single-step clean estimate $\hat{\mathbf{x}}_0$ at each step. Using a high-quality reconstruction from the same model as reference $\mathbf{x}_0^*$ (e.g., a 50-step denoised result) ensures a faithful approximation of the ideal states, allowing Diffusion DNA to better trace the model’s intended denoising trajectory. The resulting error trajectory—termed \textbf{Diffusion DNA}—compactly encodes input-specific generative difficulty dynamics.

\begin{figure}[t]
   
  \centering

  \includegraphics[width=\linewidth]{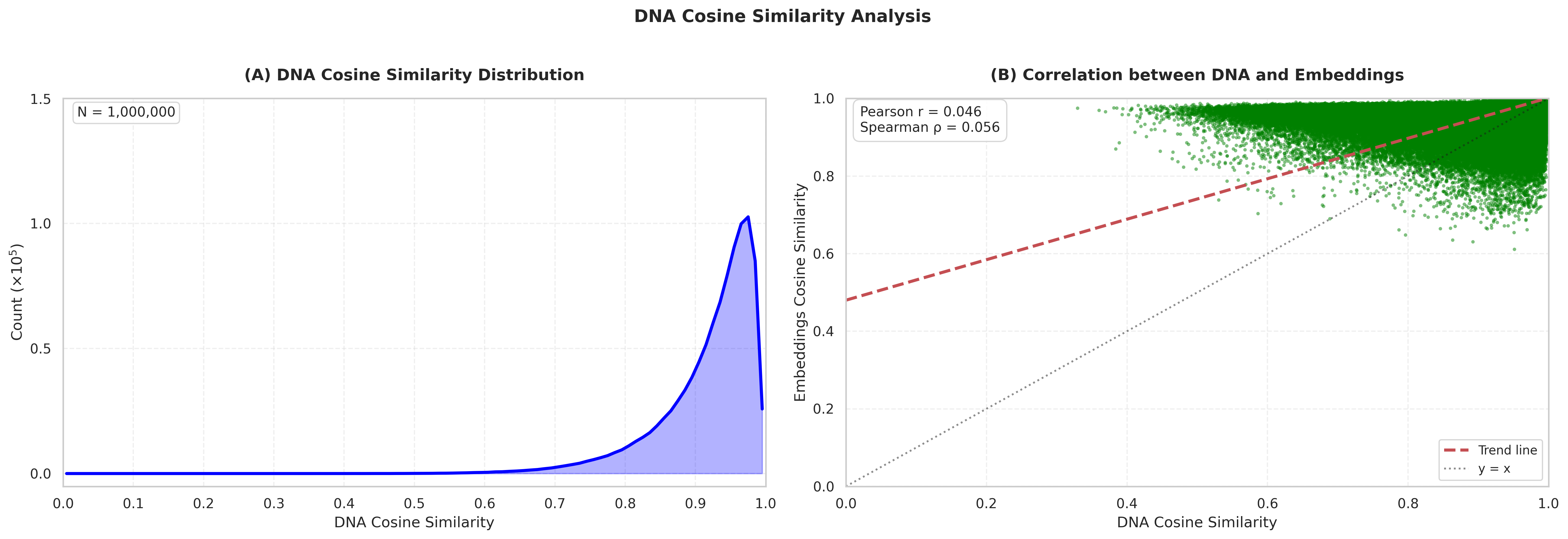} 

\caption{
\textbf{Statistical landscape of Diffusion DNA.}
We analyze large-scale structural properties of difficulty profiles
using 1,000,000 prompt pairs derived from the PickScore training set.
(A) \textbf{DNA cosine similarity distribution:}
pairwise similarities concentrate in a high-similarity regime,
suggesting that diffusion difficulty profiles inhabit a structured
region of the generative landscape.
(B) \textbf{Semantic–difficulty relationship:}
the correlation between semantic embedding similarity and DNA similarity
is low ($r = 0.046$), revealing a statistical decoupling between
surface-level prompt semantics and intrinsic generative difficulty.
}

\label{fig:dna_cosine}

\end{figure}

\paragraph{\textbf{Entropic Divergence in Generative Difficulty.}}
As illustrated in Fig.~\ref{fig:dna_divergence}, the profile of $\mathcal{C}(t)$ reveals marked heterogeneity across inputs. Visually constrained scenes with low structural entropy—such as dark-toned, coherent compositions—exhibit rapid error decay, reflecting strong semantic determinacy. In contrast, abstract and visually uncertain compositions display prolonged refinement phases. The shaded region in the figure highlights the additional computational burden induced by entropic complexity, confirming that generative difficulty is primarily governed by intrinsic visual uncertainty rather than surface-level prompt properties (e.g., length or linguistic complexity). This variability motivates adaptive allocation of computational effort along the diffusion trajectory.

\paragraph{\textbf{Computational Cost and the Need for Prediction.}}
Although Diffusion DNA provides a powerful descriptor of task difficulty, its direct acquisition is computationally expensive, requiring a full reference denoising process and multiple $\mathbf{x}_0$ estimations. As shown in Sec.~\ref{sec:method_dna}, DNA can be mapped onto a graph for planning, enabling resource-aware trajectory optimization. To make adaptive scheduling practical, we therefore seek a surrogate that can predict DNA from readily available conditions, ideally from the prompt alone. This naturally raises the question: can generative difficulty be directly inferred from semantic content?

\paragraph{\textbf{Statistical Decoupling of Semantics and Difficulty.}}
To assess whether generative difficulty can indeed be inferred from semantic content, we conducted a large-scale analysis on 1,000,000 prompt pairs randomly sampled from the PickScore training set (25,432 unique prompts) \cite{kirstain2023pick}. Each prompt's DNA vector was computed as above, and we examined two distributions (Fig.~\ref{fig:dna_cosine}). First, pairwise cosine similarities among DNA vectors (Fig.~\ref{fig:dna_cosine}A) concentrate in a high-similarity regime, indicating that difficulty profiles are not arbitrary but reside in a structured region of the generative landscape. Second, and more importantly, the Pearson correlation between semantic embedding similarity and DNA similarity is strikingly low ($r = 0.046$, Fig.~\ref{fig:dna_cosine}B). This statistical decoupling shows that denoising difficulty arises from a deep, non-linear transformation of semantic information and cannot be inferred from surface-level prompt features alone. At the same time, the high structural similarity of Diffusion DNA across inputs suggests that intrinsic generative dynamics follow consistent, condition-dependent trajectories, implying that DNA lies on a predictable, low-dimensional manifold. Together, these observations motivate the use of a lightweight nonlinear model to map condition embeddings to DNA representations efficiently.

\begin{figure}[t]  
\centering  
\includegraphics[width=\linewidth]{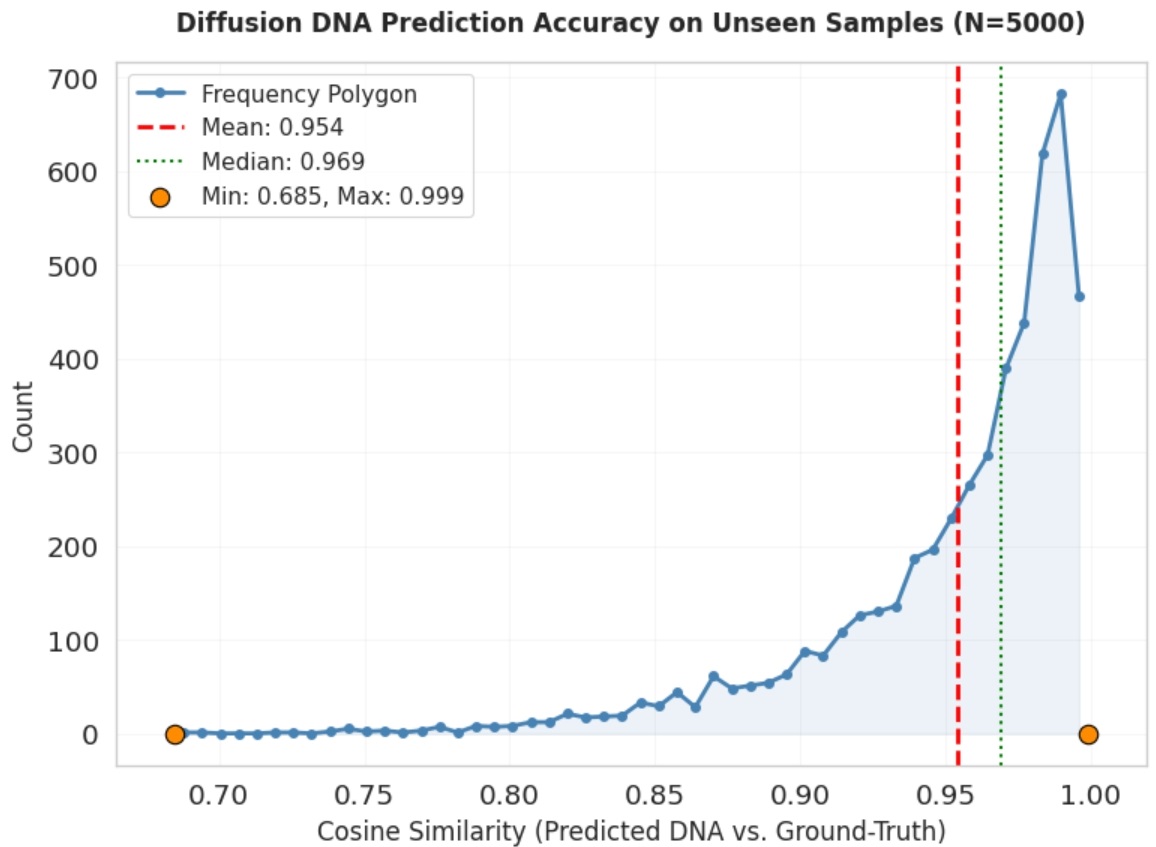}  
\caption{  
\textbf{Predicting Diffusion DNA from condition embeddings.} Distribution of pairwise cosine similarities between predicted ($\hat{\mathcal{D}}$) and true DNA trajectories. The model effectively captures structural patterns of generative difficulty across low- and high-entropy scenarios, providing a reliable signal for resource-aware, System~2 trajectory planning.  
}  
\label{fig:dna_pred}  
\end{figure}

\subsection{Predicting Diffusion DNA from Condition Embeddings}
\label{sec:dna_prediction}
\begin{table}[h]
\centering
\caption{Efficiency of the Diffusion DNA predictor.}
\label{tab:dna_efficiency}
\resizebox{\linewidth}{!}{%
\begin{tabular}{lccc}
\toprule
Model & Params (M) & FLOPs (M) & Latency (ms) \\
\midrule
DNA Predictor & 0.96 & 1.93 & 0.073 \\
\bottomrule
\end{tabular}%
}
\end{table}

\begin{table*}[t]
\centering
\caption{
\textbf{Compositional Reasoning under Fixed-Step Planning.} 
Performance comparison of CoTj (Ours) and baseline models on base and distilled models in text-to-image generation, evaluated using \textbf{GenEval}. All evaluations use a 1st-order solver, with Baseline employing Euler and CoTj (Ours) using our planned sequence. Results demonstrate that trajectory planning consistently improves reasoning performance under identical step budgets.
}
\label{tab:combined_results_fixed}
\resizebox{\linewidth}{!}{
\begin{tabular}{l c c c c c c c c}
\toprule
\textbf{Model / Method} & \textbf{Steps} & \textbf{Colors} & \textbf{Single Obj} & \textbf{Two Obj} & \textbf{Attribute} & \textbf{Counting} & \textbf{Position} & \textbf{Overall $\uparrow$} \\
\midrule
\multicolumn{9}{c}{\textbf{Pretrained Multi-Step-Based Models (not included in main comparison)}} \\
\midrule
\textcolor{gray}{SDXL \cite{Lin2024SDXL}} & \textcolor{gray}{50} & \textcolor{gray}{0.85} & \textcolor{gray}{0.98} & \textcolor{gray}{0.74} & \textcolor{gray}{0.23} & \textcolor{gray}{0.39} & \textcolor{gray}{0.15} & \textcolor{gray}{0.55} \\
\textcolor{gray}{PixArt-$\Sigma$ \cite{Chen2024PIXART}} & \textcolor{gray}{20} & \textcolor{gray}{0.80} & \textcolor{gray}{0.98} & \textcolor{gray}{0.59} & \textcolor{gray}{0.15} & \textcolor{gray}{0.50} & \textcolor{gray}{0.10} & \textcolor{gray}{0.52} \\
\textcolor{gray}{SD3-Medium \cite{esser2024scaling}} & \textcolor{gray}{28} & \textcolor{gray}{0.67} & \textcolor{gray}{0.98} & \textcolor{gray}{0.74} & \textcolor{gray}{0.36} & \textcolor{gray}{0.63} & \textcolor{gray}{0.34} & \textcolor{gray}{0.62} \\
\textcolor{gray}{FLUX-Dev \cite{flux2024}} & \textcolor{gray}{50} & \textcolor{gray}{0.79} & \textcolor{gray}{0.98} & \textcolor{gray}{0.81} & \textcolor{gray}{0.35} & \textcolor{gray}{0.74} & \textcolor{gray}{0.22} & \textcolor{gray}{0.66} \\
\textcolor{gray}{Lumina-Image-2.0 \cite{qin2025lumina}} & \textcolor{gray}{18} & \textcolor{gray}{-} & \textcolor{gray}{-} & \textcolor{gray}{0.87} & \textcolor{gray}{0.67} & \textcolor{gray}{0.62} & \textcolor{gray}{-} & \textcolor{gray}{0.73} \\
\textcolor{gray}{SANA-0.6B \cite{xie2025sana}} & \textcolor{gray}{20} & \textcolor{gray}{0.88} & \textcolor{gray}{0.99} & \textcolor{gray}{0.76} & \textcolor{gray}{0.39} & \textcolor{gray}{0.64} & \textcolor{gray}{0.18} & \textcolor{gray}{0.64} \\
\textcolor{gray}{SANA-1.6B \cite{xie2025sana}} & \textcolor{gray}{20} & \textcolor{gray}{0.88} & \textcolor{gray}{0.99} & \textcolor{gray}{0.77} & \textcolor{gray}{0.47} & \textcolor{gray}{0.62} & \textcolor{gray}{0.21} & \textcolor{gray}{0.66} \\
\midrule
\multicolumn{9}{c}{\textbf{Base Model: Qwen-Image}} \\
\midrule
\rowcolor{gray!10} Qwen-Image (Baseline) \cite{Wu2025QwenImage} & 50 & 0.90 & 0.99 & 0.98 & 0.69 & 0.89 & 0.65 & 0.85 \\
\rowcolor{gray!10} Qwen-Image (Baseline) \cite{Wu2025QwenImage} & 10 & 0.82 & 0.93 & 0.66 & 0.46 & 0.77 & 0.54 & 0.70 \\
\rowcolor{gray!10} \textbf{CoTj (Ours)} & 50 & 0.90 & 0.99 & 0.98 & 0.72 & 0.91 & 0.65 & 0.88 \\
\rowcolor{gray!10} \textbf{CoTj (Ours)} & 10 & 0.91 & 0.99 & 0.97 & 0.70 & 0.88 & 0.65 & 0.85 \\
\midrule
\multicolumn{9}{c}{\textbf{Distilled Model: Z-Image-Turbo}} \\
\midrule
\rowcolor{gray!10} Z-Image-Turbo (Baseline) \cite{Cai2025zimage} & 8 & 0.86 & 1.00 & 0.92 & 0.63 & 0.72 & 0.49 & 0.77 \\
\rowcolor{gray!10} Z-Image-Turbo (Baseline) \cite{Cai2025zimage} & 4 & 0.82 & 0.93 & 0.66 & 0.46 & 0.77 & 0.54 & 0.70 \\
\rowcolor{gray!10} Z-Image-Turbo (Baseline) \cite{Cai2025zimage} & 2 & 0.79 & 0.98 & 0.47 & 0.32 & 0.57 & 0.32 & 0.58 \\
\rowcolor{gray!10} \textbf{CoTj (Ours)} & 4 & 0.86 & 1.00 & 0.91 & 0.64 & 0.74 & 0.50 & 0.78 \\
\rowcolor{gray!10} \textbf{CoTj (Ours)} & 2 & 0.87 & 1.00 & 0.89 & 0.61 & 0.73 & 0.49 & 0.77 \\
\bottomrule
\end{tabular}
}
\end{table*}
Building on Sec.~\ref{sec:dna_discovery}, we train a lightweight three-layer MLP (ReLU, dropout) on 25,432 embedding--DNA pairs from the PickScore training set using a cosine loss (Sec.~\ref{sec:ptp}), where condition embeddings are obtained from the base model’s prompt encoder. As shown in Fig.~\ref{fig:dna_pred}, the predicted DNA trajectories closely match the ground-truth trajectories on previously unseen prompts, achieving a mean cosine similarity of 0.954 and a median of 0.969. The predictor maintains strong structural alignment across diverse and heterogeneous semantic conditions, demonstrating that the underlying generative dynamics can be reliably anticipated from conditioning signals.

The predictor is computationally lightweight (Tab.~\ref{tab:dna_efficiency}), containing 0.96M parameters and requiring 1.93M FLOPs, with an average inference latency of 0.073\,ms per prompt. This negligible overhead enables real-time estimation of Diffusion DNA prior to sampling, allowing intrinsic generative dynamics to be amortized into a predictive signal that directly supports subsequent resource-aware, System~2 planning in the Path of Least Action analysis.

\begin{figure*}[t]
\centering
\includegraphics[width=\linewidth]{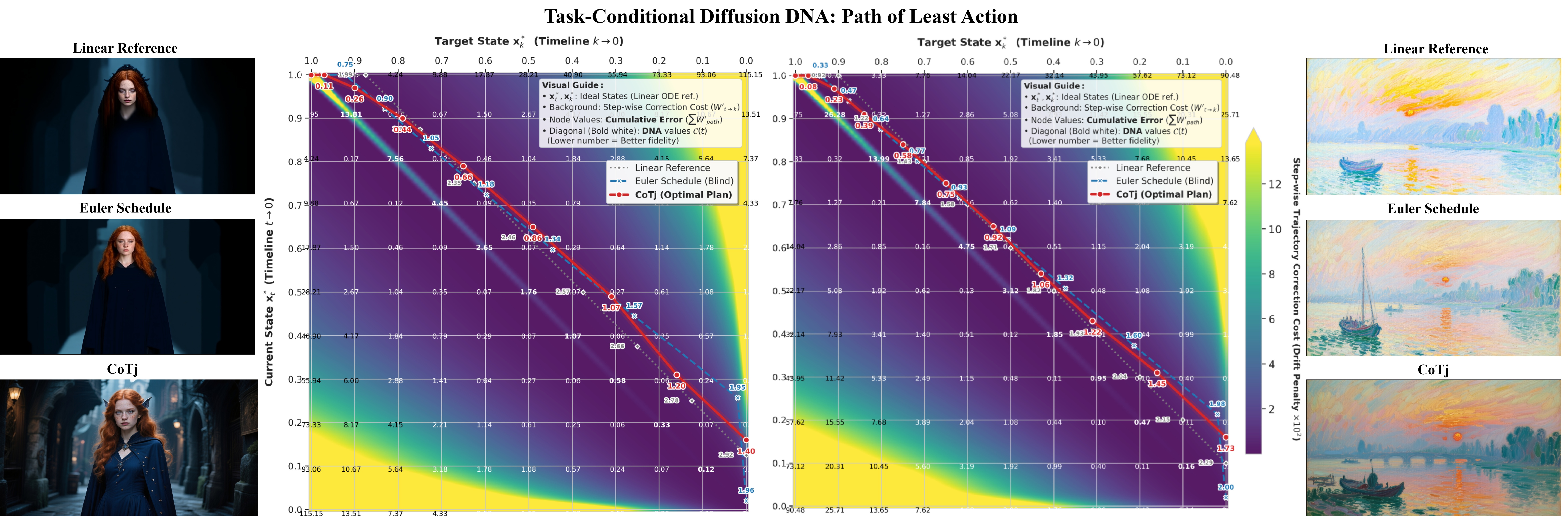}
\caption{
\textbf{Navigating the Path of Least Action.} 
\textbf{(Center)} Transition cost heatmap visualizing drift penalties and correction costs from Diffusion DNA. 
\textbf{(Left)} Low-entropy prompt: CoTj-selected trajectory of 8 steps for visualization. 
\textbf{(Right)} High-entropy prompt: CoTj-selected trajectory of 10 steps for visualization. 
All baseline trajectories use the same number of steps as the CoTj-selected path for fair comparison.
}
\label{fig:path_least_action}
\end{figure*}

\subsection{Systematic Trajectory Planning from Predicted Diffusion DNA}

Building on the predicted Diffusion DNA (Sec.~\ref{sec:dna_prediction}), CoTj constructs a transition graph (Sec.~\ref{sec:method_dna2graph})
and performs deterministic trajectory planning prior to sampling. We evaluate CoTj under two operational regimes: fixed-step optimization and adaptive-length deliberation (Sec.~\ref{sec:method_planing}), highlighting its System~2-like allocation of computational effort.

\paragraph{\textbf{Fixed-Step Planning.}}  
We evaluate CoTj under a fixed sampling budget, where timestep allocation is planned per prompt using the predicted Diffusion DNA. This allows a fair comparison with standard fixed-step inference while reflecting condition-adaptive denoising dynamics.  Table~\ref{tab:combined_results_fixed} reports compositional reasoning performance of CoTj (Ours) versus baseline models in the text-to-image generation domain, evaluated on both the base model (Qwen-Image \cite{Wu2025QwenImage}) and the distilled model (Z-Image-Turbo \cite{Cai2025zimage}). All experiments use a single-stage (1st-order) solver: baseline models employ the standard Euler sampler, while CoTj follows the same number of steps along planned trajectories derived from the predicted diffusion dynamics.  

\begin{figure}[t]
\centering
\includegraphics[width=\linewidth]{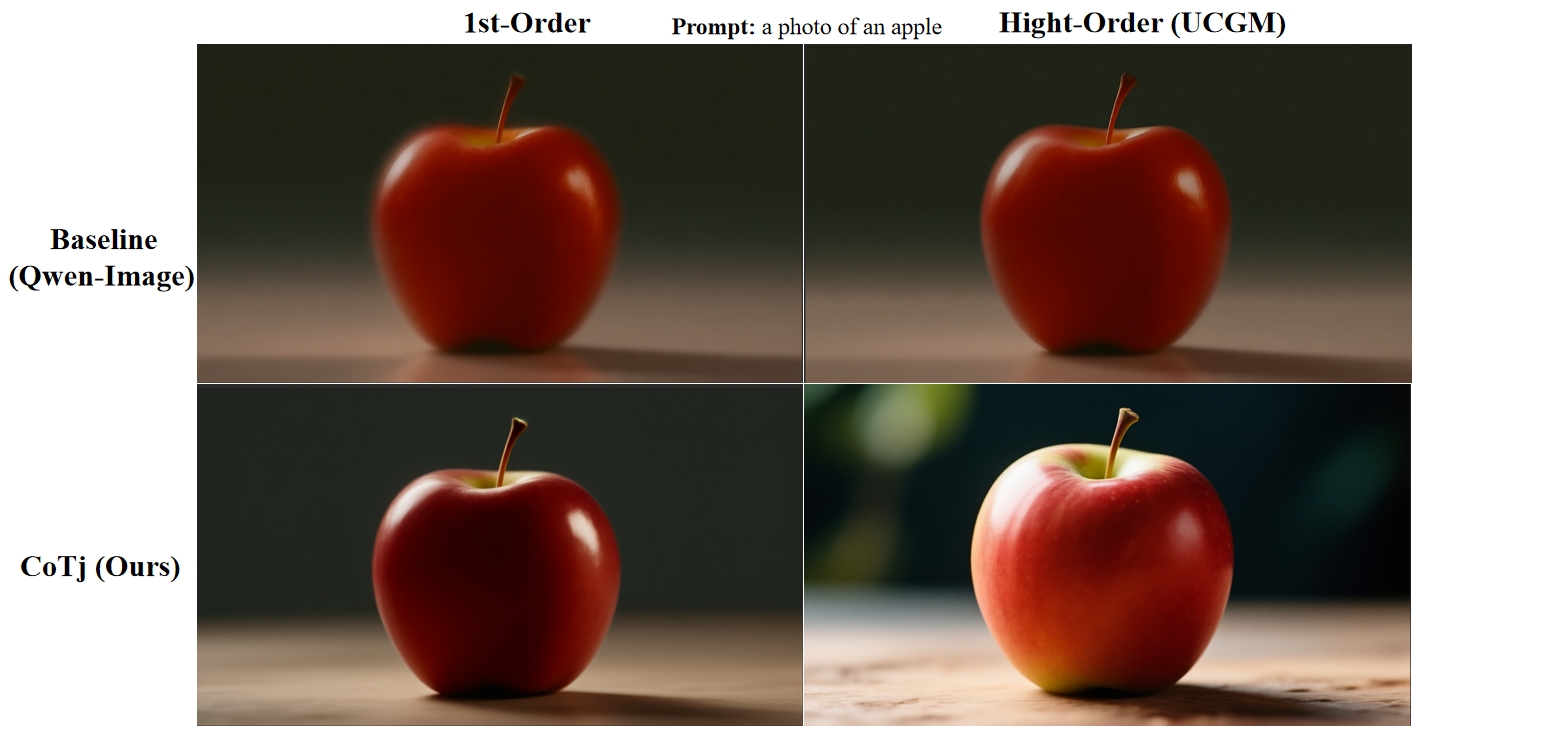} 
\caption{
\textbf{Visual Ablation at 5 Steps on Qwen-Image (Prompt: ``a photo of an apple'').}
Top-Left: Baseline shows blurred structure. Top-Right: Baseline + UCGM improves silhouette but lacks texture fidelity. Bottom-Left: CoTj (1st-Order) preserves geometry and lighting. Bottom-Right: CoTj + UCGM yields the highest realism, capturing fine textures and specular highlights.
}
\label{fig:visual_ablation}
\end{figure}

Even under extremely low-step budgets, baseline models often produce degraded outputs, whereas CoTj maintains reasonable quality (see Fig.~\ref{fig:concept} and Fig.~\ref{fig:visual_ablation}). Across both base and distilled models, CoTj consistently improves performance in all compositional reasoning metrics. 
These results highlight that conventional fixed-step schedules misallocate computation: easy regions receive redundant updates, while structurally complex stages are under-sampled. This motivates a more principled, adaptive-length strategy to explicitly model and visualize computation allocation along the denoising trajectory.

\paragraph{\textbf{Adaptive-Length Planning: Emergent System~2 Behavior.}}

While fixed-step planning improves trajectory efficiency, standard heuristics remain fundamentally limited by rigid, input-agnostic allocation of computation. Deliberative (System~2-like) reasoning, in contrast, dynamically adapts resources to task complexity. We realize this via the adaptive-length planner (Sec.~\ref{sec:method_planing}), which jointly determines step placement and total step count. Sampling terminates once the cumulative trajectory gain reaches $\rho(n)=0.99$, capturing ~99\% of the model-specific quality upper bound $W_{\max}$ (50 for Qwen-Image; for other models, set according to their recommended default steps, e.g., 8 for Z-Image-Turbo), allowing CoTj to determine minimal sufficient steps while skipping redundant evaluations. To illustrate this behavior, we visualize the predicted Diffusion DNA for representative low- and high-entropy prompts, showing how CoTj allocates computation along the minimal-cost trajectory.

\textbf{The Geometry of the Path of Least Action.}  
Figure~\ref{fig:path_least_action} shows transition cost heatmaps derived from the predicted DNA, where brighter regions correspond to higher \textit{drift penalties} or \textit{correction costs}. Standard first-order solvers such as Euler traverse this landscape using rigid schedules (dashed blue trajectories), wasting steps in low-cost plateaus while under-sampling high-cost transitions. In contrast, CoTj (solid red trajectories) identifies the minimal-cost geodesic, termed the path of least action. Under $\rho(n)=0.99$, CoTj adaptively determines minimal sufficient trajectories. For a low-entropy prompt (e.g., a dark-robed figure), only a few steps are required, skipping redundant late-stage updates. For a high-entropy prompt (e.g., an impressionist sunset), CoTj densifies nodes in critical texture-forming intervals. All baseline comparisons in this figure use the same number of steps as CoTj to ensure fair evaluation. For visualization purposes, the selected trajectories are illustrated as \textbf{8 steps} for the low-entropy prompt and \textbf{10 steps} for the high-entropy prompt. At these automatically determined step counts, CoTj preserves both structural coherence and fine detail, whereas rigid schedules often fail to resolve semantic structure or introduce severe artifacts.

\textbf{Emergent Phase Transition in Inference Dynamics.} To determine whether adaptive allocation reflects a structural property of the generative manifold rather than anecdotal prompt cases, we analyze both the spatial distribution of timesteps and the global inference dynamics across 25,432 prompts under Qwen-Image.

\begin{figure}[t]
\centering
\includegraphics[width=\linewidth]{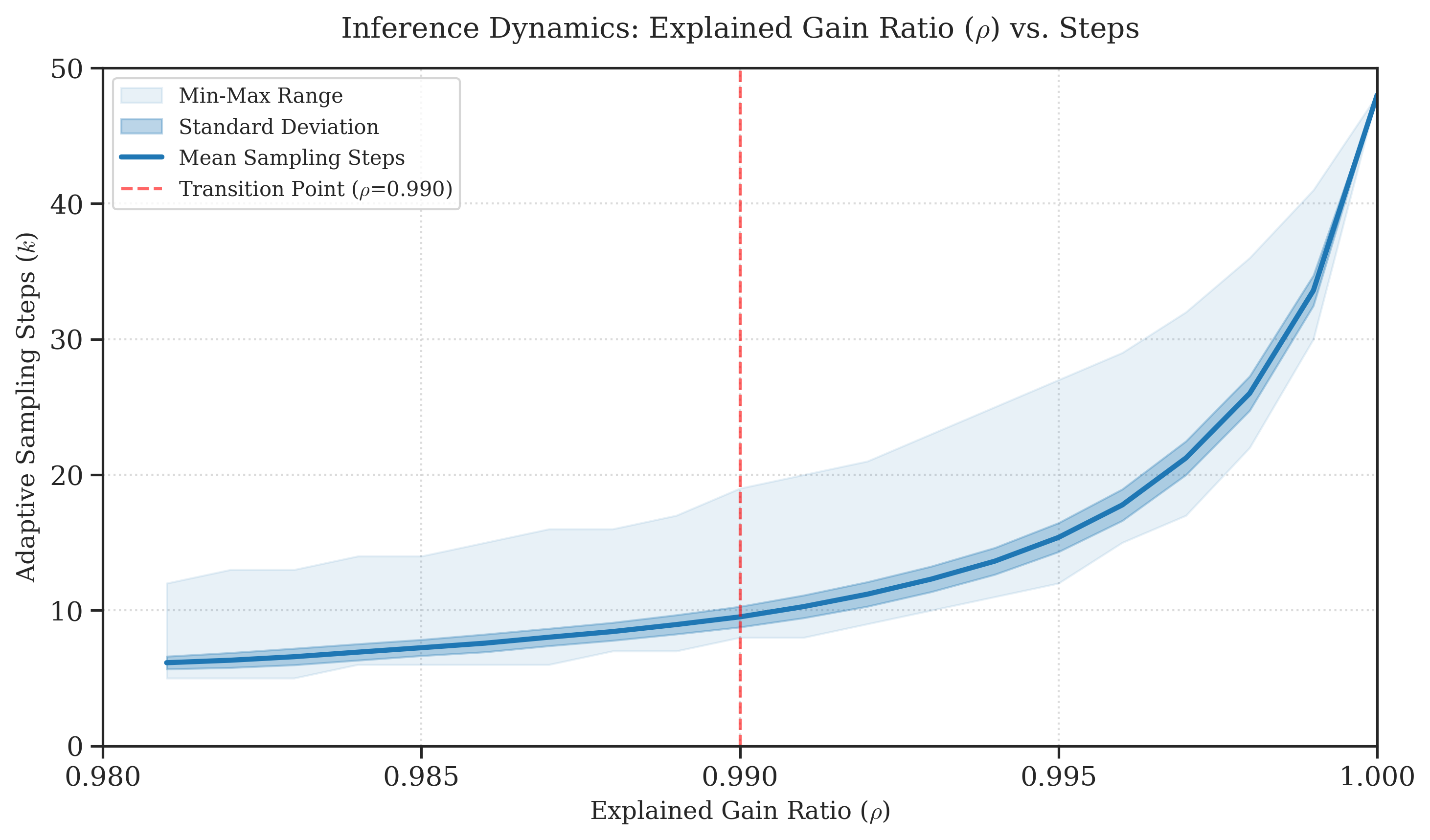} 
\caption{
\textbf{Inference Dynamics: Explained Gain Ratio ($\rho$) vs. Adaptive Sampling Steps.}
Mean adaptive step count as a function of the cumulative trajectory gain $\rho(n)$ under the Qwen-Image model. 
Shaded regions denote the min–max range and standard deviation across 25,432 prompts. 
A clear transition emerges near $\rho=0.99$ (red dashed line), beyond which required steps grow super-linearly, indicating diminishing marginal returns. 
This phase transition justifies our termination criterion, capturing near-optimal reconstruction while avoiding the exponential cost of approaching $\rho \rightarrow 1$.
}
\label{fig:gain_ration_and_adaptive_steps}
\end{figure}

Figure~\ref{fig:gain_ration_and_adaptive_steps} reveals a striking phase transition in adaptive-length planning. As the explained gain ratio $\rho(n)$ increases, the required number of steps grows gradually until $\rho \approx 0.99$, beyond which the curve rises sharply. This super-linear regime indicates diminishing marginal returns: approaching the final 1\% of trajectory gain requires disproportionately more evaluations. The shaded variance bands show that this transition is consistent across diverse prompts, confirming that it reflects an intrinsic geometric property of the diffusion manifold rather than prompt-specific noise. Consequently, $\rho=0.99$ emerges as a principled operating point, capturing near-optimal reconstruction while avoiding exponential computational cost.

\begin{figure}[ht]
\centering
\includegraphics[width=\linewidth]{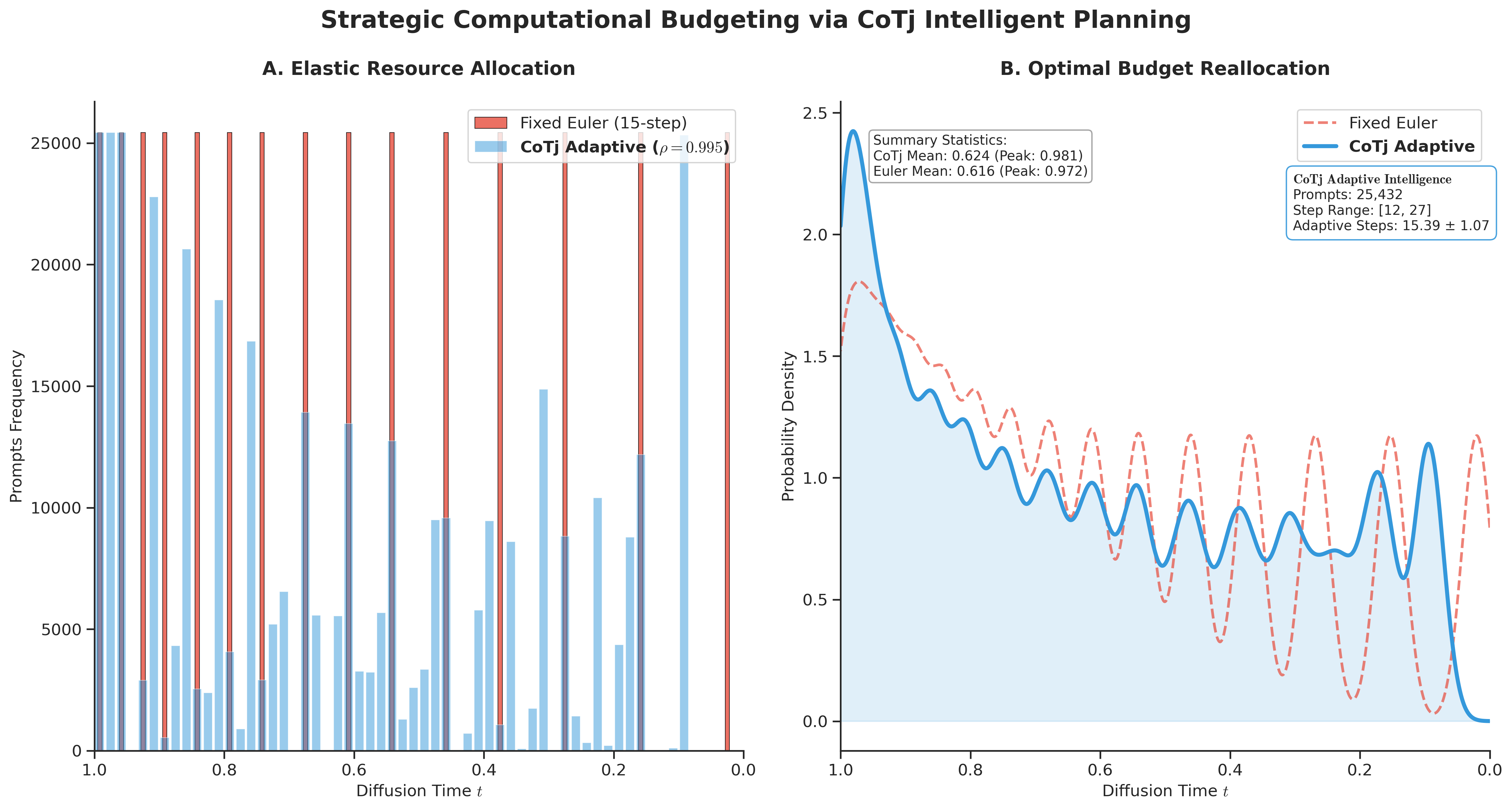}
\caption{
\textbf{System~2 Resource Allocation Across 25,432 Prompts (Qwen-Image).}  
\textbf{(A) Discrete timestep histogram:} The Euler baseline (red) uses a uniform, input-agnostic grid, whereas CoTj (blue) adaptively distributes steps, resulting in variable, prompt-specific intervals.  
\textbf{(B) Kernel Density Estimation (KDE):} The continuous density shows Euler’s uniform placement (dashed red) versus CoTj’s adaptive allocation (solid blue), concentrating computation in early high-curvature regions where global semantic structures emerge and sparsifying late low-gradient intervals.
}
\label{fig:resource_allocation}
\end{figure}

Complementing this global view, Fig.~\ref{fig:resource_allocation} analyzes how computation is distributed along diffusion time under a stricter threshold $\rho(n)=0.995$. For fair comparison, the Euler baseline is matched to the same average step count ($\approx 15$). Kernel Density Estimation reveals fundamentally different integration geometries: Euler enforces uniform step placement, implicitly assuming homogeneous reconstruction difficulty, whereas CoTj concentrates evaluations in early high-curvature regions (peak at $t=0.981$) where global semantics crystallize, while sparsifying low-gradient late intervals.

Together, these results demonstrate that fixed schedules misallocate computational resources both globally (failing to respect diminishing returns) and locally (ignoring curvature heterogeneity). By leveraging predicted Diffusion DNA, CoTj aligns solver evaluations with the intrinsic topology of the generative trajectory, transforming inference from rigid execution into deliberative, System~2-style planning.


\begin{table}[t]
\centering
\caption{
\textbf{Decoupling Trajectory Planning from Solver Complexity.}
Comparison of GenEval scores across trajectory schedules and solver types.
}
\label{tab:solver_ablation}
\resizebox{0.95\linewidth}{!}{
\begin{tabular}{llcc}
\toprule
\textbf{Method} & \textbf{Solver Type} & \textbf{Steps} & \textbf{GenEval$\uparrow$} \\
\midrule
\multicolumn{4}{l}{\textit{Few-Step Regime (Extreme Acceleration)}} \\
Baseline & Euler (1st-Order) & 5 & 0.428 \\
Baseline & UCGM (High-Order) & 5 & 0.528 \\
\textbf{CoTj (Ours)} & Planned 1st-Order & 5 & \textbf{0.626} \\
\rowcolor{gray!10} \textbf{CoTj (Ours)} & UCGM (High-Order) & 5 & \textbf{0.775} \\
\midrule
\multicolumn{4}{l}{\textit{Standard Regime (Quality Saturation)}} \\
\textbf{CoTj (Ours)} & UCGM (High-Order) & 7 & 0.811 \\
\textbf{CoTj (Ours)} & UCGM (High-Order) & 10 & 0.850 \\
\textbf{CoTj (Ours)} & Planned 1st-Order & 10 & 0.850 \\
\bottomrule
\end{tabular}
}
\end{table}

\subsection{The Primacy of Trajectory Planning}
\label{sec:ablation_solver}

While higher-order samplers have been a major focus in diffusion model research, the role of trajectory design has been less systematically explored. We investigate whether generation quality is more sensitive to the sampling path itself or to the solver applied along it. To disentangle the effects of \textit{trajectory optimization} (System 2 planning) and \textit{solver complexity} (System 1 execution), we conduct both quantitative (Table~\ref{tab:solver_ablation}) and visual analyses (Fig.~\ref{fig:visual_ablation}) on the FlowEdit dataset~\cite{kulikov2025flowedit}.

\textbf{Key Observations.} In the extreme low-step regime (Steps=5), standard linear schedules fail to preserve global structure, producing blurred and oversmoothed results. Even with UCGM \cite{Sun2025UCGM}, the Baseline cannot fully recover fine details. In contrast, CoTj with a naive 1st-order solver restores global geometry and semantic alignment, demonstrating that trajectory planning is the dominant factor. Applying UCGM on top of CoTj further enhances high-frequency fidelity, showing that planned trajectories provide an optimal substrate for higher-order corrections.

Quantitatively, CoTj with a 1st-order solver achieves a GenEval score of 0.626, surpassing Baseline + UCGM (0.528). The additional gain from combining CoTj with UCGM is +0.149 (0.775), highlighting a synergy between trajectory planning and solver order. As the step budget increases to 10, the advantage of high-order solvers becomes negligible, indicating that optimized trajectories alone can support high-quality generation without expensive solvers.

\begin{table}[t]
\centering
\caption{\textbf{Trajectory Reachability and Cache Adaptation.}  
\textbf{Top:} Planned trajectories demonstrate \textbf{Trajectory Reachability}, showing how the method follows geometrically optimal paths while preserving latent structure.  
\textbf{Bottom:} Cache comparison (\textbf{Cache Adaptation}) highlights how partial representations are efficiently reused, reducing reconstruction cost and preserving fine-grained information. Even under limited resources, relevant latent traces remain accessible, suggesting that what is present can still be recovered.}
\label{tab:reachability_cache_split}
\resizebox{\linewidth}{!}{%
\begin{tabular}{l l c c c c}
\toprule
\multicolumn{6}{c}{\textbf{Trajectory Reachability}} \\
\cmidrule(lr){1-6}
\textbf{Steps} & \textbf{Method} & MSE $\downarrow$ & SSIM $\uparrow$ & PSNR $\uparrow$ & LPIPS $\downarrow$ \\
\midrule
10 & Baseline          & 39.67 & 0.63 & 13.78 & 0.51 \\
   & CoTj$^-$       & 16.56 & 0.83 & 19.82 & 0.27 \\
   & \textbf{CoTj} & \textbf{15.36} & \textbf{0.84} & \textbf{20.36} & \textbf{0.26} \\
\midrule
17 & Baseline          & 26.11 & 0.76 & 17.08 & 0.33 \\
   & CoTj$^-$        & 7.40  & 0.92 & 25.53 & 0.14 \\
   & \textbf{CoTj}              & \textbf{6.85}  & \textbf{0.93} & \textbf{26.09} & \textbf{0.13} \\
\midrule
25 & Baseline          & 12.95 & 0.88 & 22.36 & 0.18 \\
   & CoTj$^-$        & 5.50  & 0.94 & 27.70 & 0.10 \\
   & \textbf{CoTj}              & \textbf{5.19}  & \textbf{0.95} & \textbf{27.93} & \textbf{0.09} \\
\bottomrule
\end{tabular}
}


\resizebox{\linewidth}{!}{%
\begin{tabular}{l l c c c c}
\toprule
\multicolumn{6}{c}{\textbf{Cache Adaptation}} \\
\cmidrule(lr){1-6}
\textbf{Steps} & \textbf{Method} & MSE $\downarrow$ & SSIM $\uparrow$ & PSNR $\uparrow$ & LPIPS $\downarrow$ \\
\midrule
10 & LeMiCa  \cite{gao2025lemica}  & 14.60 & 0.85 & 20.63 & 0.26 \\
   & \textbf{CoTj}             & \textbf{11.93} & \textbf{0.86} & \textbf{22.40} & \textbf{0.26} \\
\midrule
17 & LeMiCa  \cite{gao2025lemica}  & 4.81  & 0.94 & 29.23 & 0.14 \\
   & \textbf{CoTj} & \textbf{4.76}  & \textbf{0.94} & \textbf{29.22} & \textbf{0.15} \\
\midrule
25 & LeMiCa  \cite{gao2025lemica}   & 3.48  & 0.96 & 32.77 & 0.12 \\
   & \textbf{CoTj} & \textbf{2.56}  & \textbf{0.97} & \textbf{35.20} & \textbf{0.10} \\
\bottomrule
\end{tabular}
}

\end{table}


\subsection{Trajectory Reachability and Cache Adaptation}
\label{sec:trajectory_reachability}
This section evaluates whether planned trajectories can reliably reach high-fidelity latent endpoints (\textit{Trajectory Reachability}) and whether such trajectories naturally facilitate efficient state reuse (\textit{Cache Adaptation}). We first present quantitative reconstruction metrics, followed by geometric and visual analysis.

\textbf{Trajectory Reachability.} Table~\ref{tab:reachability_cache_split} (top) reports reconstruction fidelity under identical step budgets (10, 17, 25). Across all budgets, CoTj substantially outperforms the Euler baseline. At 10 steps, CoTj reduces MSE by more than 60\% and improves PSNR by over 6 dB compared to Euler. Notably, CoTj at 10 steps achieves reconstruction quality comparable to or exceeding the Euler baseline at significantly higher step counts, indicating that the baseline schedule fails to effectively exploit the model’s intrinsic denoising capacity. Even CoTj$^-$—which uses an average Diffusion DNA map without the learned predictor—already surpasses the baseline by a large margin. This demonstrates that careful geometric allocation of solver evaluations alone is sufficient to reduce ODE integration error and reach high-fidelity latent states. The full CoTj further improves performance through prompt-specific planning, confirming that trajectory-level optimization directly translates into improved reachability of the latent optimum.

\textbf{Cache Adaptation.} Table~\ref{tab:reachability_cache_split} (bottom) evaluates compatibility with cache-based acceleration under strictly controlled scheduling. To ensure fairness, both LeMiCa and CoTj operate on the same initial time grid derived from the default 50-step sequence. Specifically, we subsample the learned 100-dim Diffusion DNA using interval-2 sampling, yielding a 50-step-aligned time sequence identical to the original solver grid. Cache planning for both methods is then performed within this shared sequence, ensuring that any performance difference arises solely from trajectory ordering rather than altered time discretization.

Under identical step budgets, CoTj consistently matches or surpasses LeMiCa, particularly at higher budgets (25 steps), where it achieves significantly lower MSE and LPIPS. 
Unlike heuristic System~1 strategies that reuse fixed intervals, CoTj selects cache update nodes by following the high-information-density regions encoded in the predicted Diffusion DNA. This produces a geometrically aligned subgraph of the original trajectory, reducing redundant evaluations while preserving reconstruction fidelity.

\begin{figure*}[t]
\centering
\includegraphics[width=\linewidth]{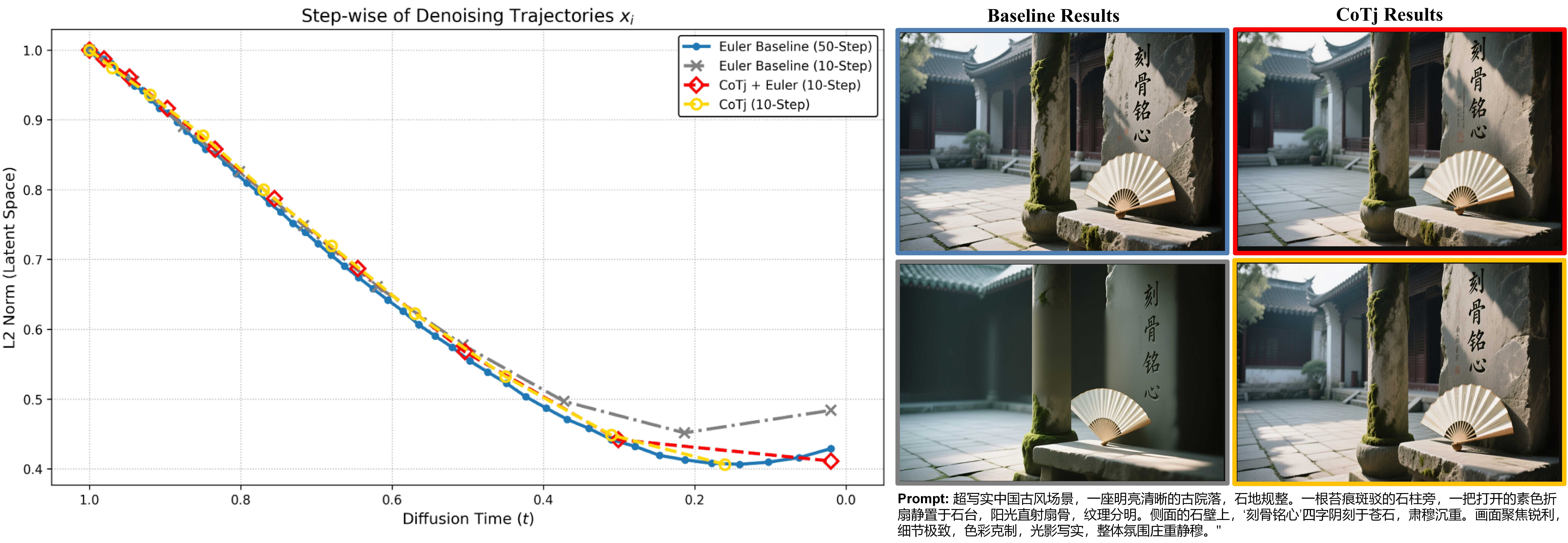} 
\caption{
\textbf{Trajectory Condensation and Structural Reachability.} 
\textbf{Left:} Step-wise L2 norm of latent states over diffusion time. Blue: Euler 50-step reference; gray dashed: Euler 10-step; red dashed: CoTj+Euler 10-step; yellow: CoTj 10-step. CoTj trajectories condense long evolutions into shorter paths while closely tracking the high-fidelity reference.
\textbf{Right:} Visual reconstructions corresponding to each trajectory (border colors match curves on the left). Within the same 10-step budget, CoTj preserves fine-grained structures lost under uniform scheduling, showing that detailed latent information can remain accessible when the correct geometric path is followed. Even when condensed, traces of what once existed persist, recoverable by mechanisms that respect the underlying structure.}
\label{fig:trajectorycondensation}
\end{figure*}

Importantly, this procedure demonstrates the universality of the learned DNA representation: the same predicted DNA can be projected onto different scheduling schemes without retraining or modifying the diffusion process. 
Moreover, the DNA learning framework is not restricted to ideal trajectories $x_t^*$; it can equivalently be trained using realized trajectories $x_t$, enabling direct adaptation to practical inference dynamics and providing a principled extension path for future cache-aware planning strategies.

\textbf{Trajectory Condensation and Geometric Alignment.} Figure~\ref{fig:trajectorycondensation} provides geometric insight into these quantitative gains. The left panel plots the L2 norm evolution of latent states. The 10-step Euler trajectory deviates significantly from the 50-step reference, exhibiting clear integration drift. In contrast, both CoTj-based trajectories (CoTj + Euler and CoTj) closely track the reference curve throughout diffusion time, effectively condensing a long high-fidelity trajectory into a shorter path. This geometric alignment directly translates into visual reconstruction quality (right panel). Under the same 10-step budget, the Euler baseline produces severe blurring and degraded text structures, whereas CoTj preserves structural coherence, sharp edges, and illumination consistency, closely resembling the 50-step reference result. 

These results demonstrate that the model’s generative capacity is not inherently limited by step count; rather, inefficient scheduling prevents the solver from reaching the latent optimum within a constrained budget. By identifying a geometrically coherent path, CoTj enables fewer evaluations to achieve fidelity previously attainable only with substantially more steps, effectively eliminating redundant computation.

\begin{figure*}[t]
\centering
\includegraphics[width=\linewidth]{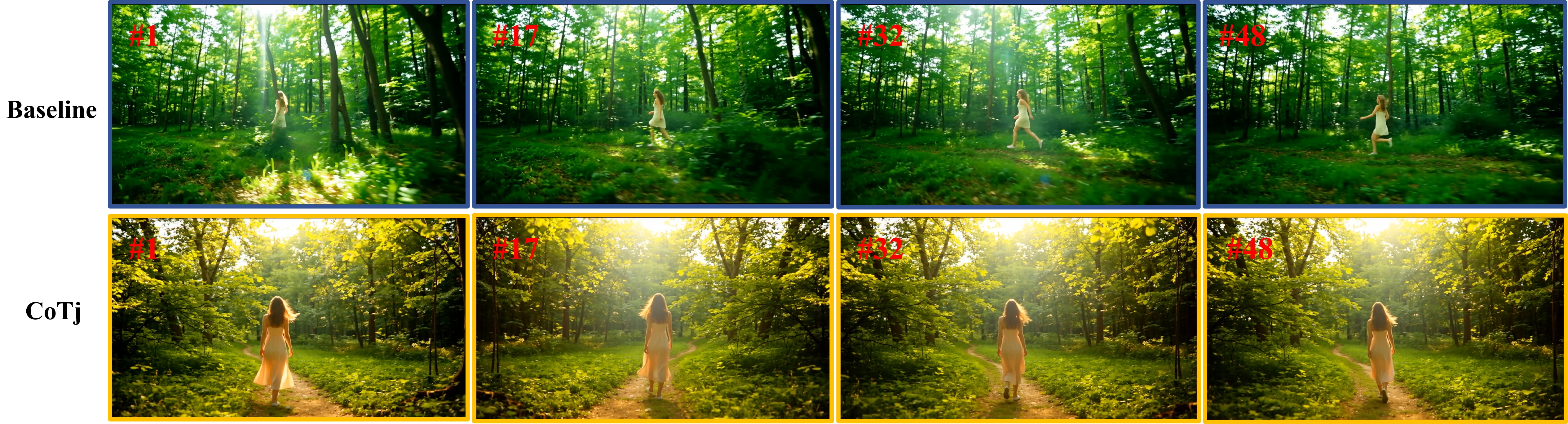}
\caption{\textbf{Qualitative comparison of 10-step structural grounding.} (Top) The baseline UniPC sampler\cite{Zhao2023UniPC} exhibits erroneous motion characterized by color shifts and structural instability beginning from the first frame. (Bottom) CoTj follows a fidelity-first planning strategy, maintaining sharp textures and stable palettes before gradually introducing motion. For the emergence of full 40-step dynamics, refer to \textbf{Fig.~\ref{fig:concept}}.}
\label{fig:video_case}
\end{figure*}

\begin{table*}[t]
\centering
\caption{VBench evaluation on Wan2.2 ($T=49$ frames). All experiments use the high-order \textbf{UniPC} sampler \cite{Zhao2023UniPC}. CoTj trajectories are planned using one-shot DNA derived from 40-step estimates. Best results are in bold.}
\label{tab:vbench_wan22}
\resizebox{\linewidth}{!}{
\begin{tabular}{l c c c c c c c}
\toprule
Method & Steps & Subject Consistency $\uparrow$ & Background Consistency $\uparrow$ & Motion Smoothness $\uparrow$ & Dynamic Degree $\uparrow$ & Aesthetic Quality $\uparrow$ & Imaging Quality $\uparrow$ \\
\midrule
Baseline & 40 & \textbf{0.9522} & \textbf{0.9549} & 0.9769 & 0.6460 & 0.5820 & \textbf{69.1599} \\
CoTj (Ours) & 40 & 0.9460 & 0.9528 & \textbf{0.9784} & \textbf{0.6549} & \textbf{0.5882} & 68.7761 \\
\midrule
Baseline & 10 & 0.9392 & 0.9493 & \textbf{0.9713} & \textbf{0.5877} & 0.5659 & 59.8147 \\
CoTj (Ours) & 10 & \textbf{0.9398} & \textbf{0.9511} & 0.9708 & 0.4912 & \textbf{0.5666} & \textbf{60.2905} \\
\bottomrule
\end{tabular}
}
\end{table*}

\subsection{Cross-Modal Adaptability: Video Generation}
To verify the cross-modal adaptability of the Chain-of-Trajectories (CoTj) framework, we extend our evaluation to video generation using the \textbf{Wan2.2} \cite{wan2025wan} model (generating 49 frames per sequence). Unlike the image-based experiments that utilize an MLP-based predictor, this analysis adopts a \textbf{direct one-shot DNA planning} approach. Specifically, we compute the Diffusion DNA by treating the 40-step denoising result as the reference $x_0$ and construct a deterministic planning graph accordingly.

Importantly, the one-shot DNA is derived from an official Wan2.2 \textit{Diffusers}\footnote{\url{https://huggingface.co/Wan-AI/Wan2.2-T2V-A14B-Diffusers}} example prompt describing two anthropomorphic cats boxing. The resulting planning graph is reused \emph{without any modification} to evaluate cross-prompt temporal generalization. For the quantitative VBench evaluation reported in Table~\ref{tab:vbench_wan22}, we follow the standard benchmark protocol. Separately, for qualitative visualization (Figure~\ref{fig:video_case}), we directly apply the same planning graph to a completely different \textbf{photorealistic} scenario: ``a young long-haired woman walking in a sunlit forest.'' This transfer from a stylized dynamic interaction to a realistic natural scene probes whether the constructed trajectory captures intrinsic model-level temporal allocation patterns rather than prompt-specific semantics. The stable behavior observed across both benchmark metrics and qualitative visualization suggests strong structural generalization of the planned trajectory. Both the baseline and CoTj-planned trajectories are executed with the \textbf{UniPC} sampler \cite{Zhao2023UniPC}, the officially recommended high-order numerical solver for Wan2.2.

\textbf{Quantitative Performance and Manifold Stability.} Table~\ref{tab:vbench_wan22} summarizes results across spatial-temporal consistency and imaging fidelity metrics. CoTj maintains high Subject and Background Consistency across both computational budgets, with negligible differences ($\leq 0.006$) compared to the baseline, indicating stable preservation of the spatial manifold. Under the extreme 10-step constraint, CoTj achieves superior \textbf{Imaging Quality} and \textbf{Aesthetic Quality}. Although the baseline reports slightly higher \textbf{Dynamic Degree} at this budget, qualitative inspection indicates that this elevated motion intensity partially arises from structural instability rather than coherent dynamics. When the sampling budget increases to 40 steps, CoTj effectively translates its structural stability into improved \textbf{Motion Smoothness} and \textbf{Dynamic Degree}, demonstrating that DNA-guided planning allocates computation more effectively to resolve complex temporal dependencies once the spatial manifold is stabilized.

\begin{figure*}[t]
\centering
\includegraphics[width=\linewidth]{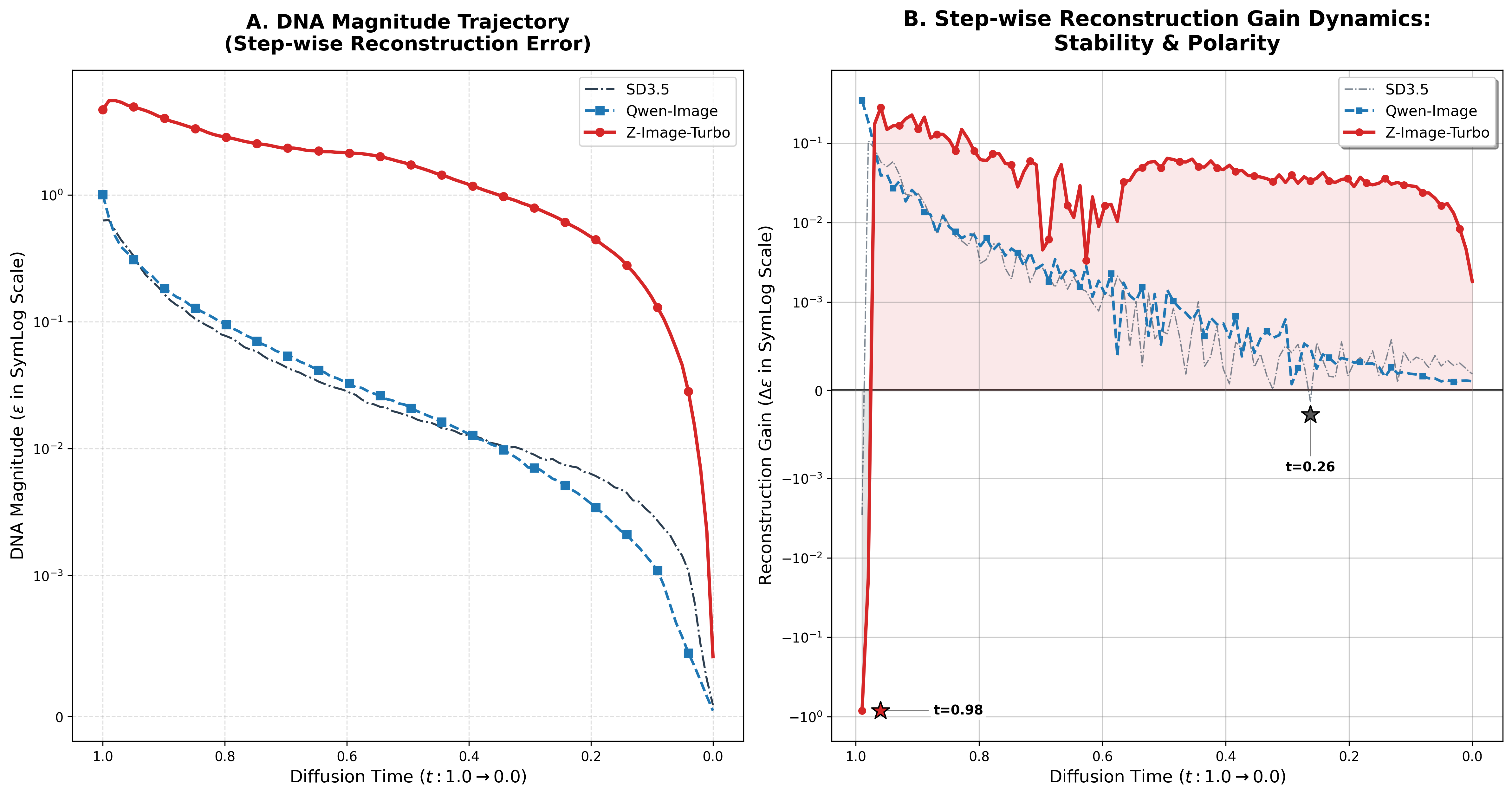}
\caption{
\textbf{Radiographic analysis of denoising dynamics.}
Curves show the dataset-averaged Diffusion DNA trajectories computed over the PickScore training set (25,432 unique prompts), ensuring consistent cross-model comparison.
\textbf{(Left)} DNA magnitude trajectory, representing absolute step-wise reconstruction error ($\varepsilon$) across diffusion time.
\textbf{(Right)} Step-wise reconstruction gain, isolating the temporal derivative of denoising progress.
Qwen-Image exhibits strictly positive and monotonically decaying gain, indicative of stable convergence. 
SD3.5 shows late-stage oscillatory behavior near zero gain, reflecting corrective inefficiencies.
Z-Image-Turbo displays a pronounced negative gain at $t=1.0$ and persistently high late-stage gain (red shaded region), revealing structural non-convergence and explaining its susceptibility to over-cooking under extended sampling. The initial negative gain further highlights why the optimal starting point for denoising may not coincide with $t=1$ in SDE-based models, motivating the Super-DAG planning design (Sec.~\ref{sec:method_dna2graph}).
}
\label{fig:dna_diagnostics}
\end{figure*}

\textbf{The Generative Hierarchy: Fidelity as the Foundation for Motion.} The primary challenge of video diffusion lies in modeling coherent temporal dynamics without compromising spatial integrity. Our findings reveal a fundamental \textbf{Generative Hierarchy}: basic attributes such as background consistency and static imaging quality are relatively easy to satisfy under limited steps, whereas high-fidelity motion requires iterative refinement across a stabilized manifold. As illustrated in Figure~\ref{fig:video_case}, the baseline at 10 steps produces visually active sequences that exhibit structural instability. Inspection of the first frame (\#1) already reveals color shifts and texture inconsistencies, indicating that temporal artifacts originate from early manifold distortion. The elevated dynamic degree in this regime therefore partly reflects sampling-induced instability—an instance of erroneous or ``pseudo'' motion rather than genuine dynamic modeling. In contrast, CoTj adopts a fidelity-first planning strategy. Under the same 10-step constraint, it prioritizes structural grounding and imaging stability, producing visually coherent videos with controlled motion amplitude. Although the measured dynamic degree is lower, the motion emerges from a preserved spatial foundation rather than instability. With an increased budget of 40 steps, CoTj leverages this stabilized structure to generate smoother and more realistic motion. As shown in Fig.~\ref{fig:concept}, CoTj ultimately surpasses the baseline in both motion smoothness and dynamic realism while maintaining strong aesthetic quality. This progression confirms that accurate motion modeling depends on prior stabilization of the spatial manifold.

These observations motivate future extensions of the DNA-to-graph formulation that explicitly incorporate temporal-entropy or motion-resistance signatures. By jointly modeling static fidelity and dynamic complexity, such extensions may further accelerate video generation while preserving coherent and physically plausible motion.

\subsection{Structural Diagnostics of Denoising Dynamics}
\label{sec:diagnostics}

To validate Diffusion DNA as a structural diagnostic for error dynamics, we perform a comparative analysis across three representative architectures: \textit{SD3.5 medium} \cite{stabilityai2024sd3}, \textit{Qwen-Image} \cite{Wu2025QwenImage}, and the distilled few-step variant \textit{Z-Image-Turbo} \cite{Cai2025zimage}. Rather than evaluating only final image quality, Diffusion DNA exposes the internal temporal structure of denoising by decomposing trajectories into magnitude and step-wise reconstruction gain. This transforms diffusion from a black-box execution process into a measurable dynamical system whose stability properties can be explicitly compared across models.

\textbf{Monotonic Gain as the Ideal Baseline.} Qwen-Image exhibits near-ideal denoising behavior. Its reconstruction gain remains strictly positive throughout the diffusion process and follows a clear monotonic decay (Fig.~\ref{fig:dna_diagnostics}). Gain is concentrated in the high-noise regime ($t = 1.0 \rightarrow 0.7$), consistent with the physical intuition that larger noise levels permit greater recoverable signal per step. As diffusion proceeds, the gain naturally diminishes, indicating progressive convergence toward the data manifold. Two structural properties follow. First, the absence of negative gain implies no denoising regression at any stage; each step contributes constructively to reconstruction. Second, the front-loaded distribution of gain implies that informational density is temporally concentrated. Consequently, step-length compression does not disrupt structural recovery, explaining the empirical robustness of base models under aggressive step reduction. Monotonic gain therefore constitutes a computable signature of stable denoising dynamics.

\textbf{Negative Gain and the Failure of Fixed Schedules.} In contrast, both SD3.5 and Z-Image-Turbo exhibit regions of negative gain, directly challenging the implicit assumption underlying uniform fixed-step schedules—namely, that each step contributes positively to reconstruction. For SD3.5, the gain oscillates around zero during late-stage diffusion ($t < 0.4$). These oscillations indicate local corrective behavior in which successive steps partially undo preceding updates. Computational effort is thus expended without consistent improvement in global fidelity. Fixed schedules are unable to discriminate these low-yield or regressive regions and therefore allocate resources suboptimally. Z-Image-Turbo exhibits a more severe pathology: a pronounced negative gain at the initial timestep ($t = 1.0$). This indicates substantial misalignment at the pure-noise boundary, introducing reconstruction error that subsequent steps must correct. Notably, this observation illustrates that the optimal denoising starting point does not always coincide with $t=1$, particularly in SDE-based models such as DDPMs\cite{Ho2020Denoising}. This observation provides structural justification for the Super-DAG design in Section~\ref{sec:method_dna2graph}, which explicitly accounts for initial regressive regions to ensure high-fidelity planning.

\textbf{Non-Convergent Gain and the Over-Cooking Phenomenon.} Beyond localized instabilities, distilled models reveal a deeper structural distinction. Unlike base models, which exhibit a natural decay of gain as $t \rightarrow 0$, Z-Image-Turbo maintains persistently high reconstruction gain even in late-stage diffusion (Fig.~\ref{fig:dna_diagnostics}, red region). This lack of gain decay indicates the absence of an inherent convergence mechanism. Whereas base models transition from structural recovery to fine-grained refinement, the distilled variant continues to perform high-energy updates throughout the trajectory. The resulting behavior explains the empirical ``over-cooking'' effect: when allowed excessive steps, the model repeatedly perturbs already-formed structures, causing deviation from the data manifold rather than convergence toward it. In this sense, distilled models are not merely unstable; they are structurally non-convergent. Diffusion DNA makes this distinction explicit by revealing the absence of a decaying gain kernel.

\textbf{From Diagnosis to Planning.} By exposing monotonicity, negative-gain regions, and convergence structure as computable quantities, Diffusion DNA enables principled trajectory planning. Our CoTj planner leverages these diagnostics to avoid regressive steps and truncate non-convergent tails, converting high-dimensional stochastic evolution into resource-aware decision-making. Diffusion DNA therefore serves not only as an interpretability tool but as a structural foundation for adaptive diffusion planning.

\section{Conclusion}
Current diffusion models operate largely in a reflexive, System~1 manner, relying on fixed, content-agnostic sampling schedules that misallocate computation and compromise fidelity, especially in high-dimensional noise manifolds. We introduce \textbf{Chain-of-Trajectories (CoTj)}, a train-free framework that equips diffusion models with deliberative, System~2-style planning. Central to CoTj is \textbf{Diffusion DNA}, a low-dimensional signature of per-step denoising difficulty that enables graph-based trajectory planning, allowing computational effort to focus on the most challenging generative phases.

Extensive experiments across image and video generation show that CoTj discovers context-aware trajectories that improve quality, structural fidelity, and motion coherence while reducing redundant computation. Diffusion DNA further provides diagnostic insight, exposing structural instabilities in distilled or few-step models and guiding adaptive planning. By decoupling \textit{what} to compute from \textit{when} to compute it, CoTj demonstrates that explicit planning can be realized in high-dimensional continuous spaces, paving the way for resource-aware, interpretable, and adaptive generative modeling. Future work includes extending CoTj to complex video dynamics, integrating online feedback for trajectory correction, and exploring unsupervised Diffusion DNA discovery across modalities.
{
    \small
    \bibliographystyle{ieeenat_fullname}
    \bibliography{main}
}


\clearpage
\setcounter{page}{1}
\onecolumn

\centering {\textbf{\Large Chain-of-Trajectories: Unlocking the Intrinsic Generative Optimality of Diffusion Models via Graph-Theoretic Planning (Appendix)}}
\raggedright
\vspace{10pt}
\appendix

\section{Theoretical Foundations of Chain-of-Trajectories}

\subsection{Proof of Postulate 1: The Upper Bound of Correctability}
\label{sec:postulate1}

\textbf{Statement:} For a canonical state $\mathbf{x}_t^*$ at timestep $t$, the single-step reconstruction error $\mathcal{C}(t)$ represents the intrinsic error upper bound for any \textit{admissible} optimal denoising trajectory originating from $t$. The goal of trajectory planning is to monotonically reduce the remaining error below this initial bound.

\textbf{Proof:} In continuous-time generative models, the reverse process is governed by either an Ordinary Differential Equation (ODE, e.g., Probability Flow) or a Stochastic Differential Equation (SDE, encompassing both drift and diffusion terms).

At any timestep $t$, the single-step reconstruction estimate $\hat{\mathbf{x}}_0(\mathbf{x}_t^*, t)$ conceptually equates to solving the reverse process from $t \to 0$ using a single, maximum-step-size projection along the predicted deterministic vector field. We define the expected error of this trivial, single-step projection as the baseline reconstruction error:
\begin{equation}
\mathcal{C}(t) \equiv \mathcal{C}(\mathbf{x}_t^*, t) = \mathbb{E}_{\mathbf{x}_0, \mathbf{z}}\bigl[\|\hat{\mathbf{x}}_0(\mathbf{x}_t^*, t) - \mathbf{x}_0\|^2\bigr]
\end{equation}

Because the true data manifold and the associated vector field $\mathbf{v}_\theta$ (or score function $\nabla_{\mathbf{x}} \log p_t$) are highly non-linear, this linear single-step projection ignores the local curvature of the probability flow, accumulating a massive global truncation error.

To formalize why $\mathcal{C}(t)$ serves as an upper bound, we consider a rational trajectory planning framework. Let $\mathcal{P}_t$ be the set of all possible multi-step discrete integration paths (using either ODE solvers or SDE ancestral sampling) from $t$ to $0$. The trivial single-step path $P_{\text{trivial}} = \{t \to 0\}$ is always a valid candidate in $\mathcal{P}_t$, with a known error of $\mathcal{C}(t)$.

If an arbitrary multi-step path $P \in \mathcal{P}_t$ utilizes unstable solver steps or poor intermediate states such that its accumulated error exceeds $\mathcal{C}(t)$, an optimal trajectory planner will simply reject $P$ and default to $P_{\text{trivial}}$. Therefore, the error of the \textit{optimally planned path} $P^*$ is strictly bounded above by the trivial baseline:
\begin{equation}
\mathbb{E}\bigl[\|\mathbf{x}_0^{(P^*)} - \mathbf{x}_0\|^2\bigr] \le \mathbb{E}\bigl[\|\mathbf{x}_0^{(P_{\text{trivial}})} - \mathbf{x}_0\|^2\bigr] = \mathcal{C}(t)
\end{equation}

Furthermore, from the perspective of numerical ODE/SDE integration, assuming the learned drift/score is locally Lipschitz continuous, refining the integration grid (i.e., introducing intermediate denoising steps $k \in (0, t)$) provides a piecewise approximation that better hugs the true probability flow. We define an \textit{admissible denoising path} as any sequence of solver steps that satisfies the region of absolute stability, thus strictly reducing the global truncation error compared to the maximum-step-size projection.

For any such admissible path, the single-step error $\mathcal{C}(t)$ intrinsically defines the maximum initial ``error budget.'' The introduction of discrete solver steps in Chain-of-Trajectories acts to systematically pay off transition costs while reducing the final reconstruction error strictly below this $\mathcal{C}(t)$ bound.

\subsection{Proof of Postulate 2: The Minimum Error Bound of the Canonical Ideal State}
\label{sec:postulate2}

\textbf{Statement:} For any timestep $t$, the reconstruction error upper bound of the ideal canonical state $\mathbf{x}_t^*$ is strictly less than or equal to that of an off-manifold state $\mathbf{x}_t$ generated via practical numerical integration. That is, $\mathcal{C}(\mathbf{x}_t^*, t) \le \mathcal{C}(\mathbf{x}_t, t)$. For notational simplicity, we denote $\mathcal{C}(t) \equiv \mathcal{C}(\mathbf{x}_t^*, t)$.

\textbf{Proof:} In both ODE and SDE formulations, the neural network (whether predicting noise $\boldsymbol{\epsilon}_\theta$, velocity $\mathbf{v}_\theta$, or directly $\hat{\mathbf{x}}_0$) is optimized via empirical risk minimization strictly over the true forward marginal distribution $q_t(\mathbf{x}_t^*)$. The training objective natively minimizes the expected reconstruction error conditioned on the ideal manifold:
\begin{equation}
\mathcal{C}(t) = \min_\theta \mathbb{E}_{\mathbf{x}_0 \sim p_{\text{data}}, \mathbf{x}_t^* \sim q_t(\mathbf{x}_t^* | \mathbf{x}_0)} \bigl[ \|\hat{\mathbf{x}}_0(\mathbf{x}_t^*, t) - \mathbf{x}_0\|^2 \bigr]
\end{equation}

During practical sampling, the actual state $\mathbf{x}_t$ is obtained by accumulating discrete numerical integration steps from $T$ down to $t$. Due to inherent truncation errors in the solver, $\mathbf{x}_t$ drifts away from the canonical data manifold $q_t$. We can express this off-manifold state as a geometric perturbation of the ideal state:
\begin{equation}
\mathbf{x}_t = \mathbf{x}_t^* + \boldsymbol{\delta}_t
\end{equation}
where $\boldsymbol{\delta}_t$ represents the cumulative off-manifold drift vector. Because $\mathbf{x}_t$ deviates from $q_t$, it acts as an Out-of-Distribution (OOD) sample for the network.

Assume the learned denoiser function $\hat{\mathbf{x}}_0(\cdot, t)$ is locally Lipschitz continuous with respect to its input state. Using a first-order Taylor expansion around the ideal state $\mathbf{x}_t^*$, the reconstruction from the off-manifold state is:
\begin{equation}
\hat{\mathbf{x}}_0(\mathbf{x}_t, t) \approx \hat{\mathbf{x}}_0(\mathbf{x}_t^*, t) + \mathbf{J}_{\hat{\mathbf{x}}_0}(\mathbf{x}_t^*) \boldsymbol{\delta}_t
\end{equation}
where $\mathbf{J}_{\hat{\mathbf{x}}_0}$ is the Jacobian matrix of the denoiser.

The expected reconstruction error for the off-manifold state $\mathbf{x}_t$ then becomes:
\begin{equation}
\mathcal{C}(\mathbf{x}_t, t) = \mathbb{E}\bigl[\|\hat{\mathbf{x}}_0(\mathbf{x}_t, t) - \mathbf{x}_0\|^2\bigr] \approx \mathbb{E}\Bigl[\|\hat{\mathbf{x}}_0(\mathbf{x}_t^*, t) - \mathbf{x}_0 + \mathbf{J}_{\hat{\mathbf{x}}_0}(\mathbf{x}_t^*) \boldsymbol{\delta}_t\|^2\Bigr]
\end{equation}

Expanding the squared $L_2$ norm yields:
\begin{equation}
\label{eq:ood_penalty}
\mathcal{C}(\mathbf{x}_t, t) \approx \mathbb{E}\bigl[\|\hat{\mathbf{x}}_0(\mathbf{x}_t^*, t) - \mathbf{x}_0\|^2\bigr] + \mathbb{E}\bigl[\|\mathbf{J}_{\hat{\mathbf{x}}_0}(\mathbf{x}_t^*) \boldsymbol{\delta}_t\|^2\bigr] + 2\mathbb{E}\bigl[\langle \hat{\mathbf{x}}_0(\mathbf{x}_t^*, t) - \mathbf{x}_0, \mathbf{J}_{\hat{\mathbf{x}}_0}(\mathbf{x}_t^*) \boldsymbol{\delta}_t \rangle\bigr]
\end{equation}

Notice that the first term is exactly $\mathcal{C}(t)$. Furthermore, because the network is optimally trained to predict the conditional expectation $\mathbb{E}[\mathbf{x}_0 | \mathbf{x}_t^*]$, the residual error $(\hat{\mathbf{x}}_0(\mathbf{x}_t^*, t) - \mathbf{x}_0)$ is orthogonal to functions of $\mathbf{x}_t^*$ in expectation, causing the cross-term to vanish to zero.

We are left with the intrinsic on-manifold error plus a strictly positive OOD penalty term:
\begin{equation}
\mathcal{C}(\mathbf{x}_t, t) \approx \mathcal{C}(t) + \mathbb{E}\bigl[\|\mathbf{J}_{\hat{\mathbf{x}}_0}(\mathbf{x}_t^*) \boldsymbol{\delta}_t\|^2\bigr]
\end{equation}

Since the penalty term $\mathbb{E}\bigl[\|\mathbf{J}_{\hat{\mathbf{x}}_0}(\mathbf{x}_t^*) \boldsymbol{\delta}_t\|^2\bigr] \ge 0$, we definitively conclude that:
\begin{equation}
\mathcal{C}(t) \le \mathcal{C}(\mathbf{x}_t, t)
\end{equation}

This rigorously establishes that the ideal trajectory mathematically guarantees the tightest possible reconstruction error bound. It provides the fundamental justification for why our trajectory planner is willing to incur the direct trajectory correction cost $W(t,k)$ in order to conceptually project the sampling path back onto the canonical ideal manifold.

\subsection{Ideal Trajectories and the Correction Cost}

During practical sampling, an actual state $\mathbf{x}_k$ is obtained by advancing from a previous state $\mathbf{x}_t$ ($t > k$): $\mathbf{x}_k = \text{Solver}(\mathbf{x}_t, t, k)$. In contrast, the ideal state $\mathbf{x}_k^*$ is obtained directly from the forward noise marginals.

The optimal denoising trajectory should strictly adhere to the ideal noise trajectory $(\mathbf{x}_T^*, \mathbf{x}_{T-1}^*, \dots, \mathbf{x}_1^*)$. However, a numerical solver step in practical sampling from $\mathbf{x}_t^*$ to $k$ deviates from this ideal path, yielding a realized state $\mathbf{x}_k$. While different solvers (e.g., higher-order Heun, DPM-Solver) and SDE discretizations introduce varying geometric drift, we can model the primary deviation using a standard first-order numerical step (e.g., Euler for ODEs or Euler-Maruyama for SDEs), which primarily steps along the deterministic vector field:
\begin{equation}
\mathbf{x}_k \approx \mathbf{x}_t^* - (t-k) \mathbf{v}_\theta(\mathbf{x}_t^*, t)
\end{equation}

The difference between this realized state and the ideal state constitutes the \textbf{Trajectory Correction Cost} $W(t,k)$. As established in Eq.~(\ref{eq:correction_cost}) of the main text, this cost projects the off-manifold geometric drift into the dimension of reconstruction error. By analyzing this first-order solver behavior, we derive the temporal scaling factor $s(t,k)$ (detailed in Appendix \ref{sec:stk}), allowing us to approximate the general transition cost as:
\begin{equation}
W(t, k) = \|\mathbf{x}_k - \mathbf{x}_k^*\|^2 \approx s(t, k) \cdot \mathcal{C}(t)
\end{equation}

\subsection{Recursive Optimization and Corollary}
\label{sec:corollary1}
For an ideal state $\mathbf{x}_t^*$ at timestep $t$, we seek to transition to a subsequent timestep in the interval $(0, t)$. We face a fundamental choice between continuing on a realized off-manifold path or paying a cost to correct back to the ideal trajectory. The optimal error bound at $t$ can be recursively defined as a search over all possible next steps $j$ (for the uncorrected path) and $k$ (for the corrected path):
\begin{equation}
\mathcal{C}(t) \leftarrow \min \Bigl( \min_{0 \le j < t} \mathcal{C}(\mathbf{x}_j, j), \; \min_{0 \le k < t} \bigl[ \mathcal{C}(k) + W(t, k) \bigr] \Bigr)
\end{equation}

Computing the specific off-manifold penalty $\mathcal{C}(\mathbf{x}_j, j)$ requires evaluating the network at every possible drifting state across all timesteps, which is computationally intractable and triggers the curse of dimensionality. We resolve this via the following corollary.

\noindent\textbf{Corollary 1: The On-Manifold Correction Assumption.} \textit{Assume the reverse-time generative ODE/SDE exhibits intrinsic error amplification over time, and the denoiser is Lipschitz continuous. Then, there exists an ideal correction step $k \in (0, t)$ that is superior to (or equal to) the best possible off-manifold continuation from any step $j \in (0, t)$:}
\begin{equation}
\exists k \in (0, t) \text{ s.t. } \mathcal{C}(k) + W(t, k) \le \min_{0 \le j < t} \mathcal{C}(\mathbf{x}_j, j)
\end{equation}

\noindent\textbf{Proof:} Let $j^*$ be the optimal timestep that minimizes the off-manifold error bound, i.e., $j^* = \arg\min_{0 \le j < t} \mathcal{C}(\mathbf{x}_j, j)$. 

Relying on Postulate 2, the error of this specific off-manifold state $\mathbf{x}_{j^*}$ can be approximated by bounding its Out-of-Distribution (OOD) penalty distance to the nearest on-manifold state $\mathbf{x}_{j^*}^*$:
\begin{equation}
\mathcal{C}(\mathbf{x}_{j^*}, j^*) \approx \mathcal{C}(j^*) + L^2 \|\mathbf{x}_{j^*} - \mathbf{x}_{j^*}^*\|^2 = \mathcal{C}(j^*) + L^2 W(t, j^*)
\end{equation}
where $L$ is the local Lipschitz constant of the denoiser.

In both diffusion SDEs and flow matching ODEs, early discretization errors are exponentially amplified as the reverse integration proceeds toward $t=0$ due to the expansive nature of the reverse vector field. This implies the effective local Lipschitz constant with respect to trajectory divergence strictly satisfies $L \ge 1$. 

Therefore, if we were to simply set our ideal step $k$ equal to $j^*$, paying the direct geometric correction cost upfront rather than letting the network amplify the error, we obtain:
\begin{equation}
\mathcal{C}(j^*) + W(t, j^*) \le \mathcal{C}(j^*) + L^2 W(t, j^*) \approx \mathcal{C}(\mathbf{x}_{j^*}, j^*)
\end{equation}

This guarantees that there exists at least one $k$ (specifically, $k=j^*$) that satisfies the inequality. Consequently, the minimum bound found in the ideal trajectory space will always be less than or equal to the minimum bound in the off-manifold space:
\begin{equation}
\min_{0 \le k < t} \bigl[ \mathcal{C}(k) + W(t, k) \bigr] \le \min_{0 \le j < t} \mathcal{C}(\mathbf{x}_j, j)
\end{equation}

Because the global minimum is mathematically guaranteed to reside within the set of ideal trajectory corrections, the outer optimization safely simplifies to a tractable, one-dimensional Dynamic Programming problem exclusively over the ideal states:
\begin{equation}
\mathcal{C}(t) \leftarrow \min_{0 \le k < t} \bigl\{ \mathcal{C}(k) + W(t, k) \bigr\}
\end{equation}

\subsection{Global Error Reduction Gain}

This dynamic programming formulation translates directly into searching for a discrete sequence of timesteps $(t_j, t_{j-1}, \dots, t_i)$ where $t_j > t_{j-1} > \dots > t_i$. Our objective is to maximize the \textbf{Error Reduction Gain} from $t_j$ to $t_i$:
\begin{equation}
\text{Gain}(t_j \to t_i) = \mathcal{C}(t_j) - \Biggl[ \sum_{m=i}^{j-1} W(t_{m+1}, t_m) + \mathcal{C}(t_i) \Biggr]
\end{equation}

Maximizing this gain is algebraically equivalent to minimizing the Total Path Cost:
\begin{equation}
\text{Cost}(t_j \to t_i) = \Biggl( \sum_{m=i}^{j-1} W(t_{m+1}, t_m) \Biggr) + \mathcal{C}(t_i) - \mathcal{C}(t_j)
\end{equation}

For the full trajectory from $T$ to a terminal state $t_i$, the initial term $\mathcal{C}(T)$ is fixed. Thus, minimizing the path cost becomes:
\begin{equation}
\text{Cost}(T \to t_i) \propto \Biggl( \sum_{m=i}^{j-1} W(t_{m+1}, t_m) \Biggr) + \mathcal{C}(t_i)
\end{equation}

This perfectly aligns with our formulation of the Super-DAG in Section \ref{sec:method_dna2graph}, where transition edges represent $W(t_{m+1}, t_m)$ and terminal edges represent the terminal risk $\mathcal{C}(t_i)$, allowing optimal trajectory planning via a standard shortest-path search.

\section{Derivation of the Temporal Lever $s(t,k)$ under Linear Flow}
\label{sec:stk}

We derive the temporal scaling factor $s(t,k)$ used in Eq.~(\ref{eq:correction_cost}) under linear flow matching \cite{lipman2022flow,Liu2023Flow}.

Consider the linear forward trajectory
\begin{equation}
\mathbf{x}_t^* = (1-t)\mathbf{x}_0 + t \mathbf{z}, 
\quad \mathbf{z}\sim\mathcal{N}(0,I),
\end{equation}
which defines the canonical state at time $t$ obtained by linearly interpolating between data $\mathbf{x}_0$ and noise $\mathbf{z}$.
For any earlier timestep $k<t$, the ideal state is
\[
\mathbf{x}_k^* = (1-k)\mathbf{x}_0 + k\mathbf{z}.
\]

At inference time, we only observe the current state $\mathbf{x}_t^*$ and a model-predicted velocity $\mathbf{v}_t$.
Assuming locally constant velocity over $(k,t)$, backward propagation gives
\[
\mathbf{x}_k = \mathbf{x}_t^* - (t-k)\mathbf{v}_t.
\]
Subtracting the ideal state yields
\begin{equation}
\mathbf{x}_k - \mathbf{x}_k^*
= (t-k)(\mathbf{v}^* - \mathbf{v}_t),
\end{equation}
where $\mathbf{v}^*=\mathbf{z}-\mathbf{x}_0$ is the ground-truth velocity under linear flow \cite{Liu2023Flow}..

We now relate this deviation to the reconstruction error at time $t$.
Under linear flow, the clean data can be recovered from the true velocity via
$\mathbf{x}_0 = \mathbf{x}_t^* - t\mathbf{v}^*$.
Replacing $\mathbf{v}^*$ with the model prediction defines the time-$t$ estimate of $\mathbf{x}_0$:
\[
\hat{\mathbf{x}}_0 = \mathbf{x}_t^* - t\mathbf{v}_t.
\]
Rearranging gives
\[
\hat{\mathbf{x}}_0 - \mathbf{x}_0
= t(\mathbf{v}^* - \mathbf{v}_t).
\]

Substituting into the deviation expression leads to
\begin{equation}
\mathbf{x}_k - \mathbf{x}_k^*
= \frac{t-k}{t}(\hat{\mathbf{x}}_0 - \mathbf{x}_0).
\end{equation}

This shows that the deviation at any intermediate timestep $k$ is proportional to the reconstruction error of $\mathbf{x}_0$ estimated at time $t$.
Hence, all temporal deviations can be reduced to a single reference quantity
$\|\hat{\mathbf{x}}_0 - \mathbf{x}_0\|$,
which unifies the scale across timesteps and anchors trajectory correction to reconstruction quality.

Taking squared norms yields
\[
\|\mathbf{x}_k - \mathbf{x}_k^*\|^2
= \left(\frac{t-k}{t}\right)^2
\|\hat{\mathbf{x}}_0 - \mathbf{x}_0\|^2,
\]
which defines the temporal lever
\begin{equation}
s(t,k) = \left(\frac{t-k}{t}\right)^2.
\end{equation}

\noindent\textit{Remark.}
Although derived under linear flow, the same predict-then-plan argument extends to SDE-based diffusion models, where an analogous temporal scaling can be estimated from local reconstruction error.

\end{document}